%% file: main.tex
\documentclass[pmlr]{jmlr}% new name PMLR (Proceedings of Machine Learning)

\usepackage{longtable}% for long tables

 \usepackage{booktabs}
 \usepackage{microtype}
\usepackage{graphicx}
\usepackage{todonotes}
\usepackage{lipsum}
\usepackage{multirow}

\usepackage{url}
\usepackage{amsmath, amssymb}
\usepackage{bm}
\usepackage{blindtext}
\usepackage{enumitem}

\usepackage{listings}
\usepackage{inconsolata}
\usepackage{bm}
\usepackage{nicefrac}
\usepackage{subcaption}
\usepackage{pifont}% http://ctan.org/pkg/pifont
\newcommand{\cmark}{\text{\ding{51}}}%
\newcommand{\xmark}{\text{\ding{55}}}%
\usepackage{mathtools}

\definecolor{red}{rgb}{1.0,0.0,0.0}
\definecolor{mydarkblue}{rgb}{0,0.08,0.45}
\definecolor{citecol}{rgb}{0.45,0.0,0.0}
\definecolor{filcol}{rgb}{0.45,0.,0.45}
\definecolor{string}{RGB}{200, 170, 0}
\definecolor{comment}{RGB}{117, 113, 94}
\definecolor{normal}{RGB}{0, 0, 0}
\definecolor{identifier}{RGB}{166, 226, 46}
\definecolor{number}{RGB}{166, 0, 166}

\newcommand{\package}{\href{https://www.cs.cmu.edu/~chiragn/redirects/auton_survival/main.html}{\texttt{auton-survival}}}

\usepackage{hyperref}
\hypersetup{colorlinks=true,
    linkcolor=mydarkblue,
    citecolor=citecol,
    filecolor=mydarkblue,
    urlcolor=mydarkblue}

\newcommand*\rot{\rotatebox{90}}

\input{math_commands}

\usepackage[load-configurations=version-1]{siunitx} % newer version

\makeatletter
\def\set@curr@file#1{\def\@curr@file{#1}} %temp workaround for 2019 latex release
\makeatother

\jmlryear{2022}
\jmlrworkshop{Machine Learning for Healthcare}

\title[\package{}]{\package{}: an Open-Source Package for Regression, Counterfactual Estimation, Evaluation and Phenotyping with Censored Time-to-Event Data}

\author{\Name{Chirag Nagpal, Willa Potosnak and Artur Dubrawski}\\
\Email{\{chiragn, wpotosna, awd\}@andrew.cmu.edu}\\
\addr Auton Lab, School of Computer Science\\
Carnegie Mellon University\\
Pittsburgh, Pennsylvania, United States} 

\begin{document}

\maketitle

\vspace{-3em}
% \centerline{\large \textbf{\url{https://autonlab.github.io/auton-survival}}}
\centerline{\color{mydarkblue} \large \textbf{\texttt{\href{https://www.cs.cmu.edu/~chiragn/redirects/auton_survival/main.html}{autonlab.github.io/auton-survival}}}}
\vspace{1em}

\begin{abstract}

Applications of machine learning in healthcare often require working with time-to-event prediction tasks including prognostication of an adverse event, re-hospitalization, and mortality. Such outcomes are typically subject to censoring due to loss of follow up. Standard machine learning methods cannot be applied in a straightforward manner to datasets with censored outcomes. In this paper, we present \package{}, an open-source repository of tools to streamline working with censored time-to-event or survival data. \package{} includes tools for survival regression, adjustment in the presence of domain shift, counterfactual estimation, phenotyping for risk stratification, evaluation, as well as estimation of treatment effects.
Through real world case studies employing a large subset of the SEER oncology incidence data, we demonstrate the ability of \package{} to rapidly support data scientists in answering complex health and epidemiological questions.
\end{abstract}

\section*{Introduction}
\label{sec:intro}

Machine learning is being increasingly applied to address challenges across multiple areas of healthcare. Healthcare data is inherently complex and thus offers a multitude of challenges to traditional machine learning. The advent of deep neural networks has especially catalyzed interest in the use of machine learning for healthcare as these models can be used to learn nonlinear representations of complex clinical data. 

Modern machine learning approaches inherently involve reasoning and inference in terms of binary or categorical outcomes. In reality, healthcare outcomes often are continuous time-to-events, such as mortality, stroke, onset of cancer, and re-hospitalization. The challenges of working with time-to-event data are further compounded by the fact that
data typically includes individuals whose outcomes are missing, or \textit{censored} due to loss of follow up. Standard classification and regression approaches do not provide a straightforward way to deal with such censored data.

%\textcolor{red}{Modern machine learning approaches most generally involve reasoning about outcomes as binary or categorical events that indicate occurrence within a time period. In reality, healthcare outcomes, such as mortality, stroke, and hospital readmission, may be more appropriately reasoned in terms of continuous time-to-event. Another consideration for modern machine learning is how to deal with censoring, or unknown outcomes that occur due to loss of individuals during follow up. Standard approaches involving classification do not provide a straightforward way to deal with such censored data.} 

Bio-statistics and medical informatics literature has extensively dealt in methodology to deal with censored time-to-event outcomes.  However, such methodology is relatively understudied in modern machine learning. As a result, there is limited support in terms of robust, reproducible software that is equipped to handle censored data. In this paper, we present the package \package{}, a comprehensive \texttt{python} repository of user-friendly utilities for application of machine learning in the presence of censored time-to-event data. 

\newpage

\noindent \textbf{Generalizable Insights about Machine Learning in the Context of Healthcare:}

\noindent Healthcare research is replete with problems involving assessment of time-to-events. From critical care, oncology, and epidemiology to cardio-vascular, mental, and public health, problems involving analysis of time-to-event outcomes are ubiquitous. Examples include mortality, hospital discharge, and onset of an adverse clinical conditions such as stroke and cancer. The analysis of such time-to-event data, especially when subjected to censoring, requires a specialized set of tools different from standard classification or regression. 

\begin{description}[leftmargin=*]

\item[Technical Significance:]
Current popular machine learning frameworks involve modelling problems as classification or regression. Applying these existing frameworks to healthcare problems often involves discretizing time-to-event outcomes as binary classification. However, this approach neglects temporal context which could result in models with misestimation and poor generalization. While bio-statistics literature has extensively dealt with methodology and corresponding software for censored time-to-event outcomes, support support for survival outcomes through modern representation and machine learning techniques is limited. \package{} is one step towards providing support to work with time-to-event data and includes an exclusive suite of workflows that allow for a multitude of experiments from data pre-processing and regression modelling to model evaluation. Additionally, \package{} uses an API similar to \texttt{scikit-learn} \citep{pedregosa2011scikit}, making adoption easy for users already familiar with machine learning in \texttt{Python}.

\package{} includes thorough documentation of utilities as well as example code notebooks to facilitate rapid prototyping. \package{} is open source and hosted on GitHub\footnote{\href{https://github.com/autonlab/auton-survival}{\texttt{github.com/autonlab/auton-survival}}} to promote widespread use of the package for reproducible machine learning for healthcare research.

\end{description}

%\todo[inline]{TODO Item for Chirag} 

%\newpage

\section*{Related Work and Contributions}

Modelling of censored time-to-event outcomes is recently gaining attention in the machine learning community. There have been successful attempts to incorporate deep non-linear representations in the classic Cox Proportional Hazards model \citep{faraggi1995neural, katzman2018deepsurv}. Considerable attention has been given to the problem of easing the strong restrictive assumptions of the proportional hazards model involving discrete time \citep{yu2011learning, lee2018deephit}  as well as parametric \citep{ranganath2016deep, nagpal2021deep} and adversarial \citep{chapfuwa2018adversarial} approaches.  

There have been other attempts at reproducible machine learning pipelines with \texttt{Python}: the \texttt{lifelines} \citep{davidson2019lifelines} and \texttt{scikit-survival} \citep{polsterl2020scikit} packages offer support for classical standard survival regression methods and \texttt{pycox} \citep{kvamme2019time} attempts to streamline the application of \texttt{torch} based deep learning models with simpler APIs. A broad comparison of the extended functionalities offered by \package{} is in \hyperref[sec:comparison]{Appendix \ref{sec:comparison}}.

\package{} builds upon many of the software design choices of existing \texttt{Python} packages for machine learning and survival analysis \citep{davidson2019lifelines, paszke2019pytorch, polsterl2020scikit}. It offers functionality for rapid experimentation with multiple classes of survival regression models and corresponding metrics to evaluate model discriminative capability and calibration. Additionally, \package{} uniquely offers easy to use APIs for counterfactual and treatment effect estimation as well as subgroup discovery, among other utilities, to solve the following real world problems involving censored time-to-events:

\begin{description}[leftmargin=*]
    \item[Counterfactual and Treatment Effect Estimation:] Clinical decision support often requires reasoning about \textit{what if}? scenarios. In situations where outcomes maybe confounded, such inference requires estimating counterfactuals by adjusting for confounding.
 
 \item[Survival Regression with Time-Varying Covariates:]
 {Real world health data often consists of multiple time-dependent observations per individual, or time-varying covariates. \package{} has support for auto-regressive deep learning models that allow learning temporal dependencies when estimating time-to-event outcome.}
 
 \item[Subgroup and Phenotype Discovery and Evaluation:]
{Inherent heterogeneity in patient populations results in differential event incidence rates conditioned on covariates and interventions. Identifying patient subgroups with differential risk can provide insight into best practices that benefit such phenogroups.}
 
 \end{description}

\tableofcontents

\newpage

\section{Time-to-event or Survival Regression}
\label{sec:regression}

%\begin{figure}[!h]
\begin{minipage}{0.5\textwidth}
Throughout this paper, we will work with a dataset of right censored instances $\mathcal{D} := \{ (\vx_i, \vt_i, \delta_i) \}_{i=1}^{n}$ where $\vx_i$ is the set of covariates of an individual. $\vt_i$ is the time to event or censoring $\mathbb{R}^{+}$ as indicated by the indicator $\delta_i \in \{ 0, 1\}$.  Time-to-event or survival estimation problem thus reduces to estimating the conditional distribution of survival notated as
\begin{align}
    \nonumber \mathbb{E}[\bm{1}\{T>\vt\}|X&=\vx]\\ 
       \nonumber &= \mathbb{P}(T>\vt|X=\vx)\\
 &= 1 - \mathbb{P}(T\leq \vt|X=\vx).
\end{align}

\end{minipage}\hfill
\begin{minipage}{0.45\textwidth}
\centering
\includegraphics[width=0.8\textwidth]{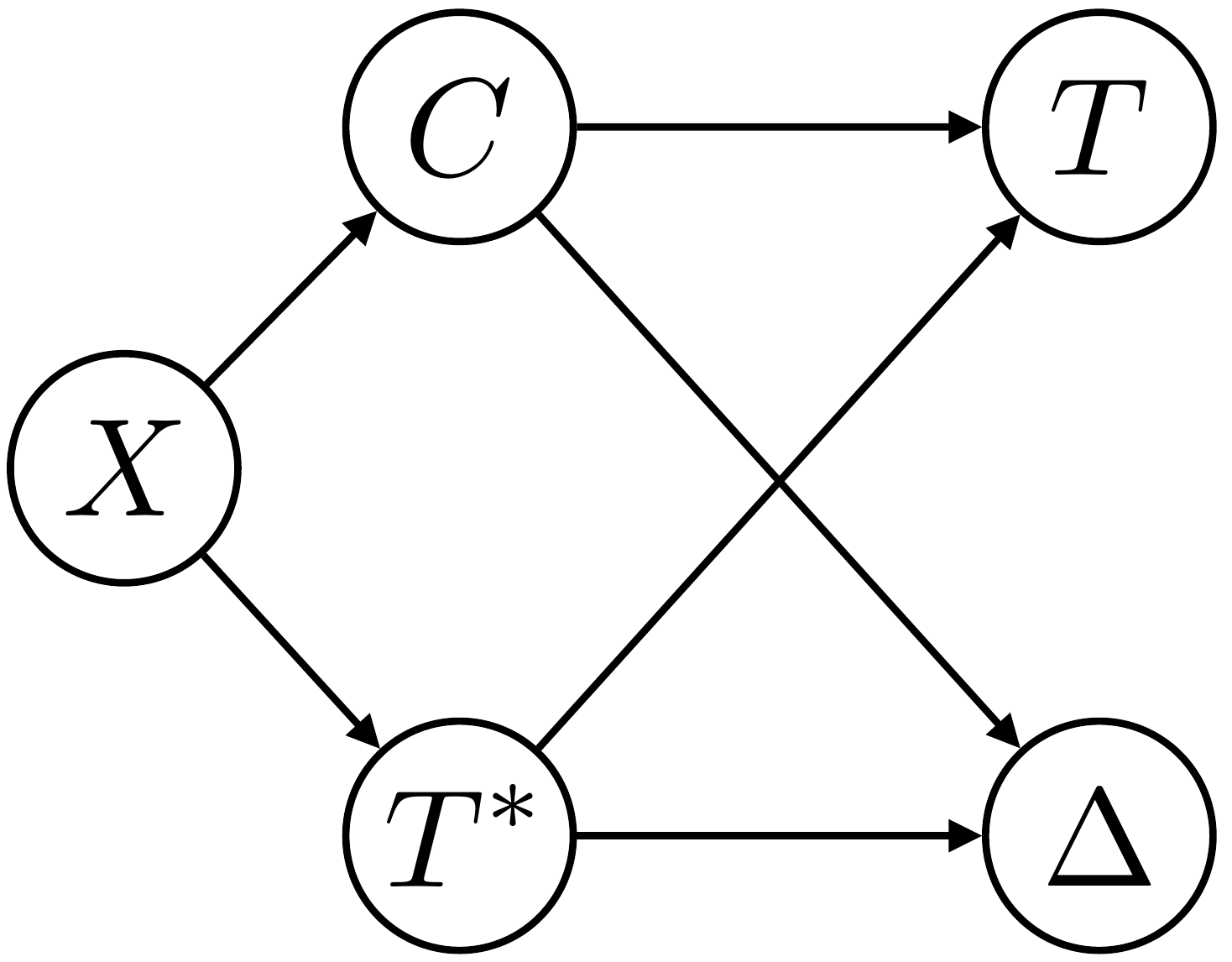}
\captionof{figure}{Conditional Independence of the True Time-to-Event $T^{*}$ and the censoring times $C$. Only $X, T$ and $\Delta$ are observed.}
\label{fig:dag-time-to-event}
\end{minipage}
%\end{figure}

%Throughout this paper we will work with a dataset of right censored instances $\mathcal{D} := \{ (\vx_i, \vt_i, \delta_i) \}_{i=1}^{n}$ where $\vx_i$ is the set of covariates of an individual. $\vt_i$ is the time to event or censoring $\mathbb{R}^{+}$ as indicated by the indicator $\delta_i \in \{ 0, 1\}$.  Time-to-event or survival estimation problem thus reduces to estimating the conditional distribution of survival notated as
\vspace{.5em}

\begin{figure}[!h]
    \centering
    \includegraphics[width=0.7\textwidth]{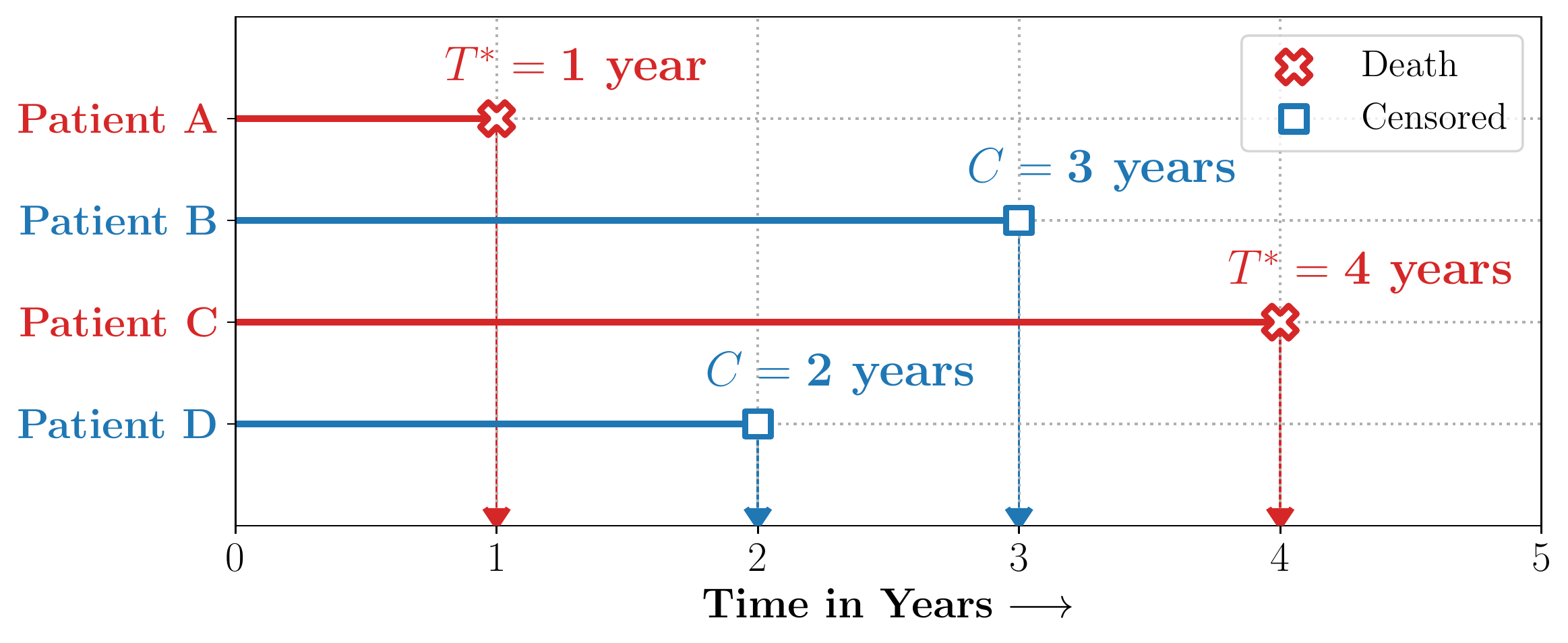}
    \caption{\textbf{Censoring and Time-to-Event Predictions}: {\color{Maroon} \textbf{Patients A}} and {\color{Maroon} \textbf{C}} died {\color{Maroon}\textbf{1}} and {\color{Maroon}\textbf{4 years}} from entry into the study, whereas {\color{MidnightBlue} \textbf{Patients B}} and {\color{MidnightBlue}\textbf{D}} exited the study without experiencing death (were lost to follow up) at {\color{MidnightBlue}\textbf{2}} and {\color{MidnightBlue}\textbf{3 years}} from entry in the study.  \textit{Time-to-Event} or \textit{Survival Regression} thus involves estimates that are adjusted for individuals whose outcomes were censored.}
    \label{fig:censoring}
\end{figure}

Here, $T$ refers to the distribution of the censored survival time $T = \textrm{min}(T^{*}, C)$, where $T^{*}$ is the distribution of the true time-to-event and $C$ is the distribution of the censoring time (\hyperref[fig:dag-time-to-event]{Figure \ref*{fig:dag-time-to-event}}). $\Delta$ is the distribution of the censoring indicator $\Delta = \bm{1}\{T^{*}<C\}$. Typically we do not observe the true event times for individuals lost to follow up as in \hyperref[fig:censoring]{Figure \ref*{fig:censoring}}.

For these individuals, we observe the censored survival time, $T$ and an indicator of if they were censored, $\Delta = T<C$. Assuming conditional independence between $T$ and $C$ ie. $T\perp C|X$ allows identification of the distribution of $\mathbb{P}(T|X)$.

For the censored individuals, we maximize the probability corresponding to the survival function. The likelihood, $\ell(\cdot)$ under censoring is thus given as
\begin{align}
    \bm{\ell}(\{ \vx, \vt, \bm{\delta}\}) \propto \mathbb{P}(T=\vt|X=\vx)^{\delta}\mathbb{P}(T>\vt |X=\vx)^{1-\delta}.
    \label{eq:likelihood}
\end{align}
Often in survival analysis literature likelihoods are expressed in terms of instantaneous hazard rates $\bm{\lambda}(t)$. The instantaneous hazard maybe defined as the event rate at a time $t$, conditional on survival $(T > t)$ till that time. Thus,
\begin{align}
    \bm{\lambda}(t) = \lim_{\textrm{d}t \rightarrow 0} \frac{\mathbb{P}(t \leq T < t+\textrm{d}t)}{\textrm{d}t\cdot S(t)} = \frac{\mathbb{P}(T=t)}{\bm{S}(t)} \quad \text{or, } \quad \bm{\lambda}(t) = - \frac{\textrm{d}}{\textrm{d}t}  \log\mS(t).
\end{align}
Now reasoning in terms of hazard rates, we can rewrite \hyperref[eq:likelihood]{Equation \ref*{eq:likelihood}} equivalently as
\begin{align}
\bm{\ell}(\{ \vx, \vt, \bm{\delta}\}) \propto    \bm{\lambda}(\vt |X=\vx)^{\delta}\mS(\vt|X=\vx).
\end{align}
Broadly the popular approaches for maximizing the likelihood above are classified as,
\begin{description}[leftmargin=*]
\item[Parametric] Assumes the distribution of the time-to-event $\mathbb{P}(T|X=\vx)$ is parametric like Weibull or Log-Normal. Examples includes the popular Accelerated Failure Time model.
\item[Non-Parametric] Involves learning kernels or similarity functions of the input covariates followed by a non-parametric (Kaplan-Meier or Nelson-Aalen) estimation of the survival rate weighted with the learnt kernel.
\item[Semi-Parametric] the Cox Proportional Hazards model and its extensions arguably, remain the most popular approaches and are classified as \textit{semi-parametric}. The Cox model involves a two step estimation where the feature interactions are learnt through a parametric model followed by non-parametric estimation of the base survival (hazard) rate. 

\end{description}

% \begin{minipage}{1\textwidth}
% \centering
% \begin{minipage}{0.5\textwidth}
% \centering
%     \end{minipage} 
%     \begin{minipage}{0.33\textwidth}
%     \centering
%     \includegraphics[width=1\textwidth]{figures/dag_time_to_event.pdf}
%     \captionof{figure}{Conditional Independence of the True Time-to-Event $T^{*}$ and the censoring times $C$.}
%     \end{minipage}
%     \label{fig:time_to_event}
% \end{minipage}

% \begin{figure}[!htbp]
%     \centering
%     \includegraphics[width=0.3\textwidth]{figures/dag_time_to_event.pdf}
%     \caption{Conditional Independence of the True Time-to-Event $T^{*}$ and the censoring times $C$. In practice we observe observe the censored survival time, $T$ and the censoring indicator, $\Delta$. Assuming conditional independence allows identification of the distribution of $\mathbb{P}(T|X)$.}
%     \label{fig:time_to_event}
% \end{figure}

\begin{figure}[!t]
    \centering
    \includegraphics[width=0.925\textwidth]{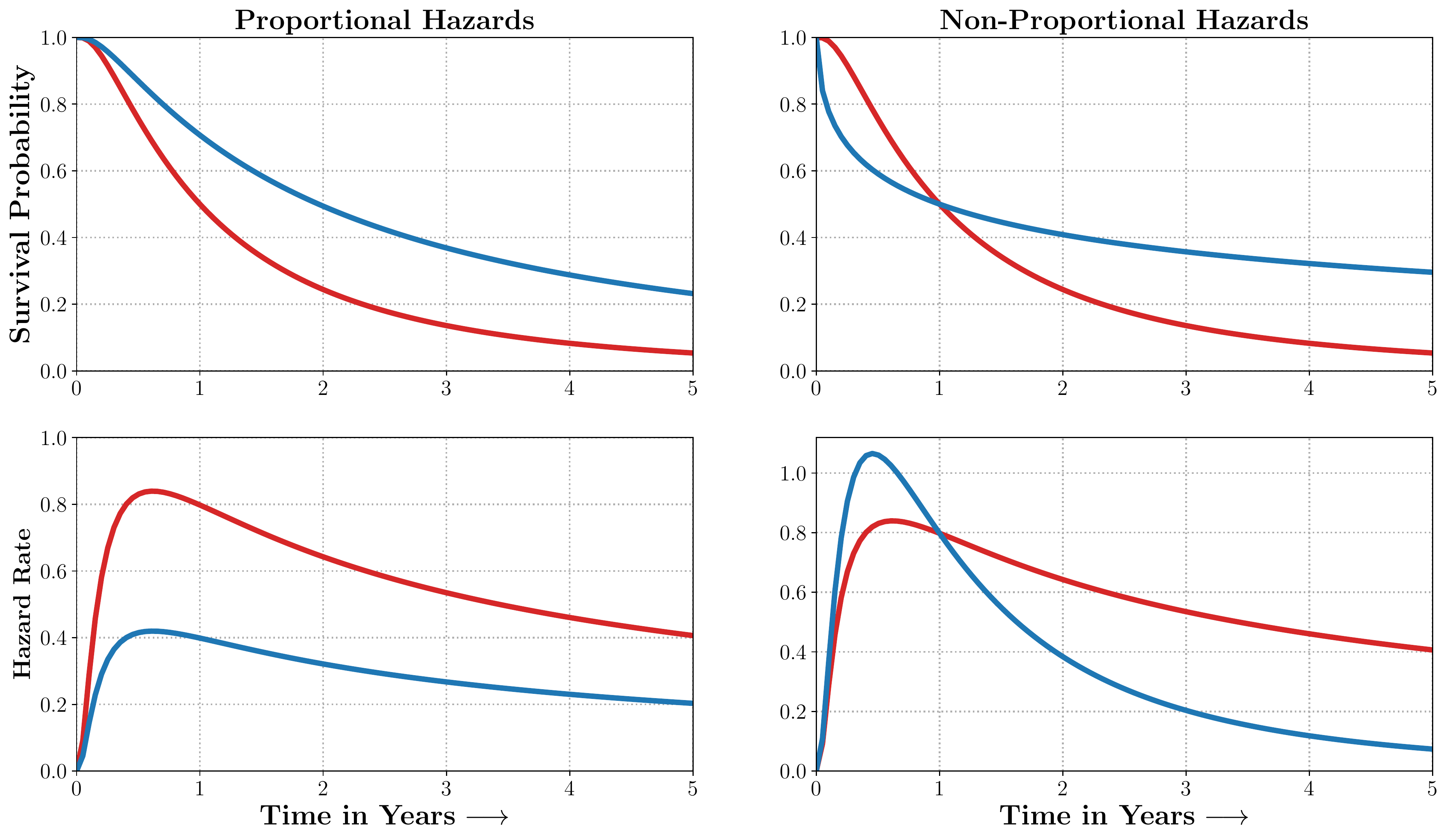}
    \caption{\textbf{Non-Proportional Hazards}: When the Proportional Hazards assumptions are satisfied, the Survival Curves and their corresponding Hazard Rates dominate each other and do not intersect. In many real world scenarios however, the survival curves. \package{} includes flexible estimators of Time-to-Events in the presence of non-proportional hazards.}
    \label{fig:ph-assumptions}
\end{figure}

\newpage 

\subsection{Fitting Survival Estimators}
\label{sec:preprocessing}

% Filled page but we can remove if too wordy
 \package{} includes the standard Cox Proportional Hazards (CPH) \citep{cox1972cph} to estimate time-to-event regression models. \package{} allows the use of deep representation learning approaches to learn the Cox model as in \cite{faraggi1995neural,katzman2018deepsurv}. Note that the standard Cox model is sensitive to the strong assumptions of \textit{Proportional Hazards} which maybe mis-specified in real world scenarios (\hyperref[fig:ph-assumptions]{Figure \ref*{fig:ph-assumptions}}). \package{} thus includes deep latent variable survival regression models \textit{Deep Cox Mixtures} (DCM) and \textit{Deep Survival Machines} (DSM) \citep{nagpal2021deep, nagpal2021dcm} that ease this restrictive assumption by modeling the conditional time-to-event distributions as mixtures of fixed size.

We demonstrate the use of \package{} to predict long term mortality for 9,105 critically ill hospitalized patients from the SUPPORT (Study to Understand Prognoses Preferences Outcomes and Risks of Treatment) \citep{supportdata, connors1995controlled} study; consisting of demographics, physiological measurements and outcomes followed up over a 5 year period. %In the following code snippet, we will load and preprocess the SUPPORT dataset using \texttt{} followed by training 

\begin{lstlisting}[escapeinside={(*}{*)}]
from auton_survival import datasets, preprocessing, models 
# Load the SUPPORT Dataset
outcomes, features = datasets.load_dataset("SUPPORT")
# Preprocess (Impute and Scale) the features
features = preprocessing.Preprocessor().fit_transform(features)
# Train a Deep Cox Proportional Hazards (DCPH) model
model = models.cph.DeepCoxPH(layers=[100])
model.fit(features, outcomes.time, outcomes.event)
# Predict risk at specific time horizons.
predictions = model.predict_risk(features, t=[8, 12, 16])
\end{lstlisting}
\vspace{.1em}

\package{} also provides the convenient, \texttt{SurvivalModel} class that enables rapid experimentation via a consistent API by wrapping multiple regression estimators. In addition to the models above, \texttt{SurvivalModel} class also includes Random Survival Forests  (RSF) \citep{ishwaran2008rsf} a popular non-parametric survival model.

\begin{lstlisting}[escapeinside={(*}{*)}]
from auton_survival import estimators (*\phantom{A} \hfill \href{https://www.cs.cmu.edu/~chiragn/redirects/auton_survival/regression.html}{{\textbf{\texttt{[Demo Notebook]}}}}*)
# Train a Deep Survival Machines model using the SurvivalModel class.
model = estimators.SurvivalModel(model='dsm')
model.fit(features, outcomes)
# Predict risk at time horizons.
predictions = model.predict_risk(features, times=[8, 12, 16])
\end{lstlisting}

% \textcolor{red}{
% \begin{description}[leftmargin=0cm]
%     \item[Cox Proportional Hazards (CPH)] \citep{cph} model assumes that individuals across the population have constant proportional hazards overtime. In this model, the estimator of the survival function is assumed to have constant proportional hazard. Thus, the relative proportional hazard between individuals is constant across time.
%     \item[Random Survival Forests (\texttt{pysurvival}):
% Weibull Accelerated Failure Time (\texttt{lifelines})] \citep{rsf} is an extension of Random Forests to the survival settings where risk scores are computed by creating Nelson-Aalen estimators in the splits induced by the Random Forest. 
% \end{description}
% }

Additionally, the \texttt{SurvivalRegressionCV} class can be used to optimize survival regression models in a $K$-fold cross validation fashion over a user specified hyperparameter grid. Model selection is performed by selecting the model that minimizes the Integrated Brier Score.

\begin{lstlisting}[escapeinside={(*}{*)}]
from auton_survival.experiments import SurvivalRegressionCV (*\phantom{A} \hfill \href{https://www.cs.cmu.edu/~chiragn/redirects/auton_survival/regression.html}{{\textbf{\texttt{[Demo Notebook]}}}}*)
# Define the Hyperparameter grid to perform Cross Validation
hyperparam_grid = {'n_estimators' : [50, 100],  'max_depth' : [3, 5],
                     'max_features' : ['sqrt', 'log2']}
# Train a RSF model with cross-validation using the SurvivalRegressionCV class
model = SurvivalRegressionCV(model='rsf', cv_folds=5, hyperparam_grid=hyperparam_grid)
model.fit(features, outcomes)
\end{lstlisting}
\vspace{.1em}

\newpage
\subsection{Importance Weighting}
\label{sec:iwerm}
Frequently in survival regression, we need to perform inference over a test dataset subject to distribution shift. Distribution shift can arise in multiple ways and affect the generalizability of a model. For simplicity, consider the covariate shift problem where,
$\mathbb{P}_{\text{train}}(T, \Delta|X) = \mathbb{P}_\text{test}(T, \Delta|X)$ but $\mathbb{P}_{\text{train}}(X) \neq \mathbb{P}_\text{test}(X)$. \package{} allows a flexible API to adjust for distribution shift  involving Importance Weighted Empirical Risk Minimization (IWERM) \citep{sugiyama2007covariate, shimodaira2000improving} by weighted resampling of the training data with replacement at training time.
\begin{align}
\mathop{\mathbb{E}}_{(\vx, \vt, \bm{\delta})\sim \mathcal{D}_\text{test}}[ \mathcal{L} (\vx, \vt, \bm{\delta}) ] &= \mathop{\mathbb{E}}_{(\vx, \vt, \bm{\delta})\sim \mathcal{D}_\text{test}}\bigg[ \ln \big( \bm{\lambda}(t|X=\vx)^{\delta} \bm{S}(t|X=\vx) \big) \bigg]\\
&= \mathop{\int}_{\vx \in \mathcal{X}}\mathop{\int}_{\vt \in \mathcal{T}}\mathop{\int}_{\delta \in \{ 0, 1\}}  \ln \bigg( \bm{\lambda}(t|X=\vx)^{\delta} \bm{S}(t|X=\vx) \bigg) \mathbb{P}_{\text{test}}(\vx, \vt, \delta)\\
&= \mathop{\mathbb{E}}_{(\vx, \vt, \bm{\delta})\sim \mathcal{D}_\text{train}}\big[\bm{w}(\vx) \cdot \ln  \big( \bm{\lambda}(t|X=\vx)^{\delta} \bm{S}(t|X=\vx) \big) \big],\\
\nonumber \text{here, }\bm{w}(\bm{x}) &\propto \frac{\mathbb{P}_\text{test}(\vx)}{\mathbb{P}_\text{train}(\vx)} \text{ are the estimated importance weights.}
\end{align}

\begin{lstlisting}
# Estimate Importance Weights with Logistic Regression
from sklearn.linear_model import LogisticRegression
p_target = LogisticRegression().fit(features, domains).predict_proba(features_source)
imp_weights = p_target/(1-p_target) # Propensity Weighting
# Train the Survival Regression Model
from auton_survival.estimators import SurvivalEstimator
model = SurvivalEstimator("dcph", layers=[100]) # Cox PH Model with 1 Hidden Layer 
model.fit(features=features_source, outcomes=outcomes_source, weights=imp_weights)
\end{lstlisting}

%\newpage

%\tcolorbox{}

%\toprule
\noindent \textbf{Cautionary Note (Censoring and Distribution Shift)} 

\noindent \begin{minipage}{\textwidth}
    \centering
    \begin{minipage}{0.5\textwidth}
    \centering
    \includegraphics[width=.7\textwidth, trim={.5cm 15cm 14cm 0cm}, clip]{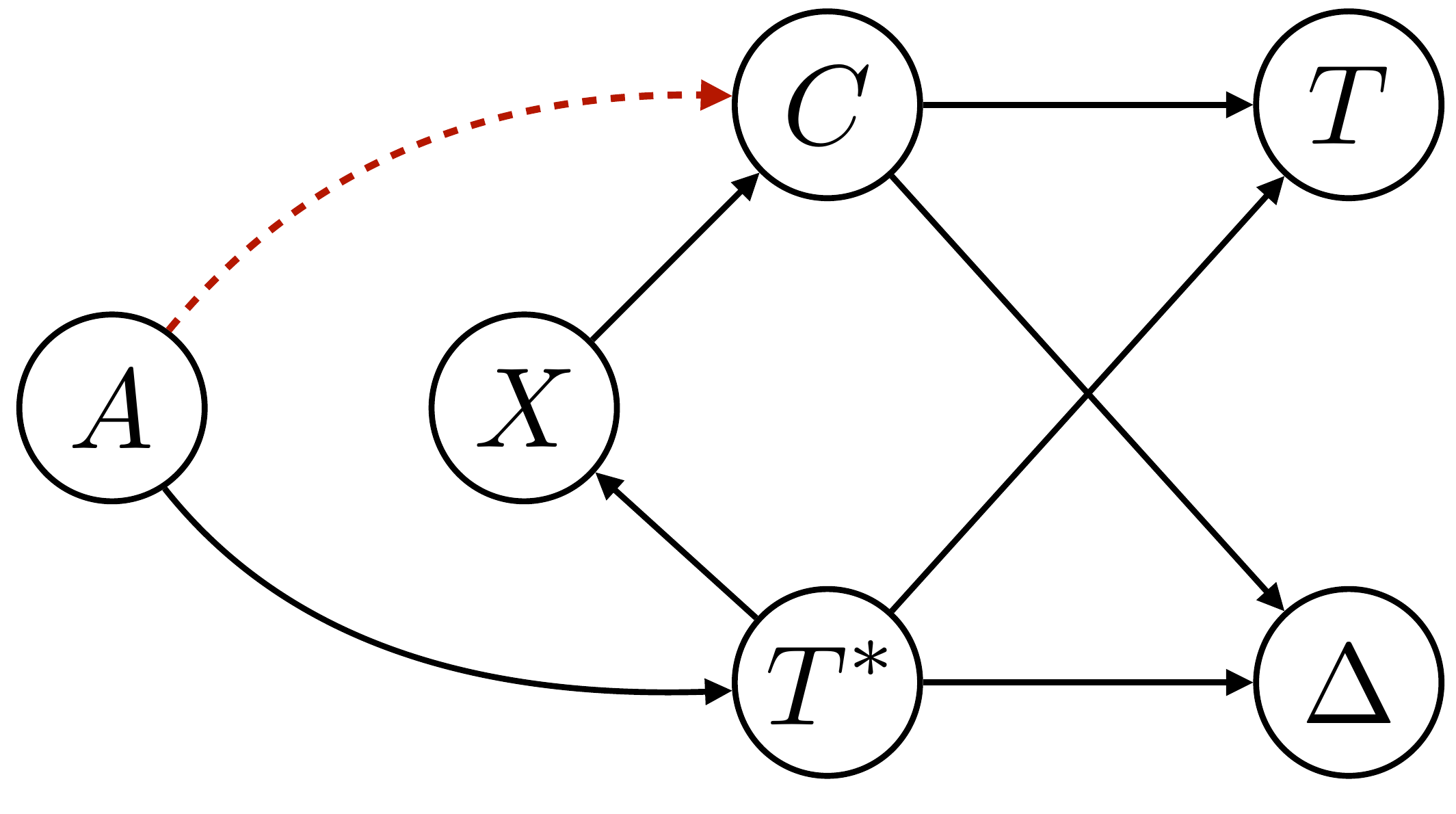}\\
    {\small \textbf{Label Shift}}
    \end{minipage}%
    \begin{minipage}{0.5\textwidth}
    \centering
    \includegraphics[width=.7\textwidth, trim={.5cm 15cm 14cm 0cm}, clip]{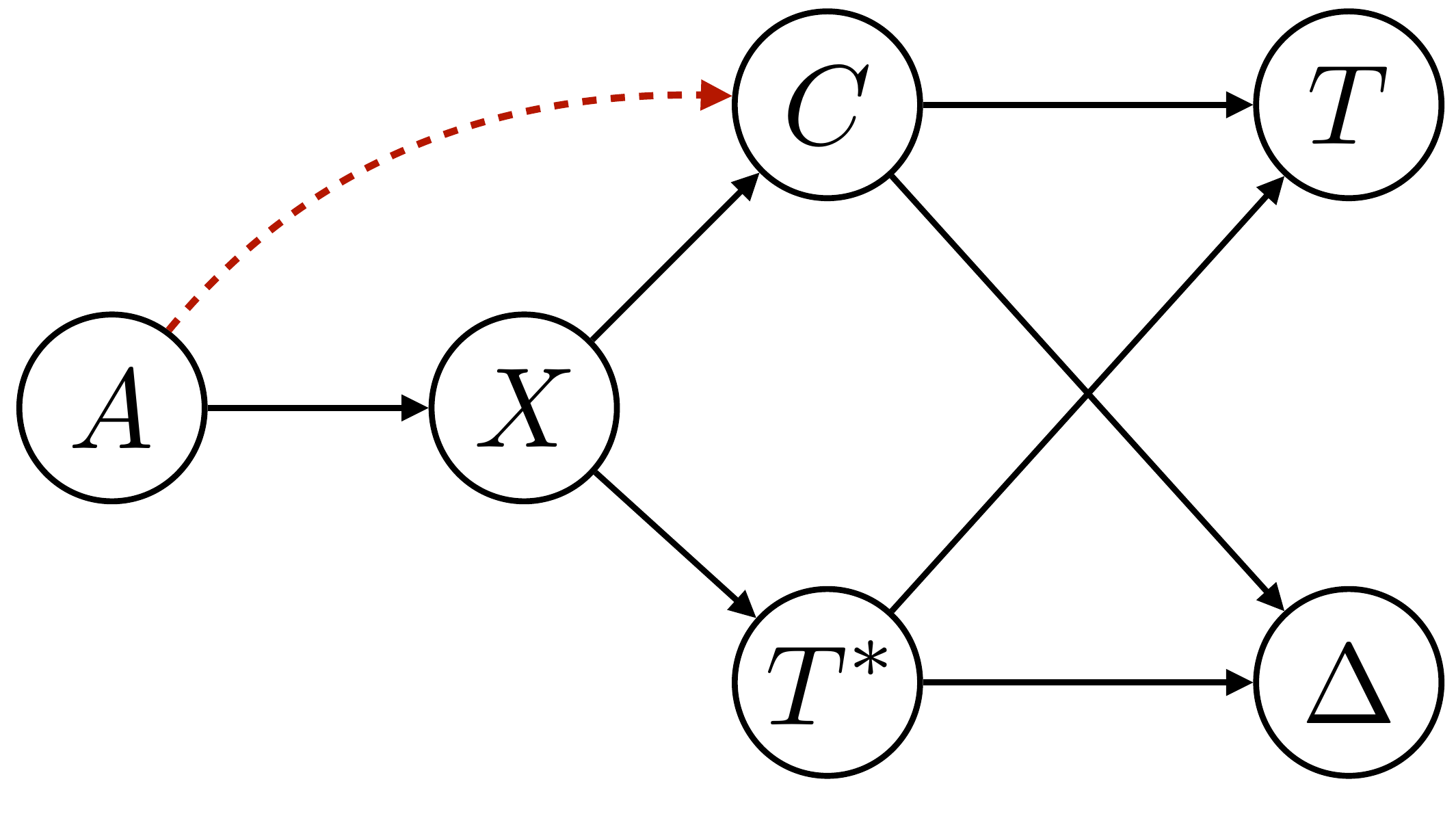}\\
    {\small \textbf{Covariate Shift}}
    \end{minipage}
    \captionof{figure}{\textit{Prior probability} and \textit{Covariate Shift} in Time-to-Event (Survival) Regression. In situations where $A \rightarrow C$ the censoring rates, $C$ are dependent on the domain $A$, it may not be possible adjust for domain shift with standard importance weighting. Readers are thus advised to exercise caution in scenarios where shifts result in such complex interactions.}
    \label{fig:time-to-event-cov}
\end{minipage}

For the above covariate shift adjustment to work, the conditional distribution of the censoring times must be the same across the source and target domains. Additionally, is common to use importance weighting for adjustment in situations where $\mathbb{P}_\textrm{train}(T) \neq \mathbb{P}_\textrm{test}(T)$ This is commonly referred to as  \textit{label shift} or \textit{prior probability shift} in Machine Learning (\hyperref[fig:time-to-event-cov]{Figure \ref*{fig:time-to-event-cov}}). As in the case of covariate shift adjustment, this requires additional assumptions on the censoring distributions which might be violated in real-world situations. Readers are thus recommended to exercise appropriate caution when reasoning about whether importance weighting is an appropriate approach to adjust for distribution shift. A thorough discussion of the implications of this is beyond the current scope of this manuscript.

\subsection{Counterfactual Survival Regression}
\label{sec:counterfactual}
Survival outcomes are often used to answer `\textit{what if?}' questions requiring inference of counterfactuals. 
We notate an intervention with an indicator $\va \in \{0, 1\}$ and assume \textbf{Strong Ignorability (Unconfoundedness)} that the potential outcomes $T(1)$ and $T(0)$ are independent of the treatment assignment, ($A=\va$) conditioned on the set of confounders ($X=\vx$) \citep{rubin2005causal}. The estimated time-to-event outcome under intervention ($A=\va$) is
\begin{align}
  \widehat{\mS}\big(t|\text{do}(A=\va)\big) =  \mathop{\mathbb{E}}_{\vx \sim \mathcal{D}}\big[ \widehat{\mathbb{E}}[\mathbf{1}\{T>t\}|X=\vx, A=\va] \big].
\end{align}
\vspace{-1em}

Note that $\widehat{\mathbb{E}}[\mathbf{1}\{T>t\}|X=\vx, A=\va]$ is just a conditional estimator of survival learnt on the population under intervention $\va$. In practice thus, the counterfactual survival regression involves fitting separate regression models on the treated and control populations. \package{} allows learning counterfactual models with $K$-fold cross-validation using the \texttt{CounterfactualSurvivalCV} class.

\begin{lstlisting}[escapeinside={(*}{*)}]
from auton_survival.experiments import CounterFactualSurvivalRegressionCV 
grid = {'layers' : [[], [100]], 'learning_rate' : [1e-3, 1e-4]} # Hyperparameter Grid
# Train a counterfactual Cox model with cross-validation.
model = CounterfactualSurvivalRegressionCV('dcph', 5, hyperparam_grid=grid)
model.fit(features, outcomes)
\end{lstlisting}

\subsection{Time-Varying Survival Regression}

\noindent \begin{minipage}{1\textwidth}
\begin{minipage}{0.48\textwidth}
Additionally \package{} also includes \textit{time-varying} implementations of Deep Survival Machines \citep{nagpal2021rdsm} and Deep Cox Proportional Hazards model \citep{lee2021leveraging} thats involves the use of \texttt{RNNs}, \texttt{LSTMs} or \texttt{GRUs} \citep{chung2014empirical, hochreiter1997long}. \hyperref[fig:tv-surv-reg]{Figure \ref{fig:tv-surv-reg}} presents time varying survival regression in \package{}. For an individual with\vspace{.4em}
\end{minipage}\hfill
\begin{minipage}{0.5\textwidth}
\centering
\includegraphics[width=1\textwidth, trim={1.5cm 10.5cm 8cm 6cm},clip]{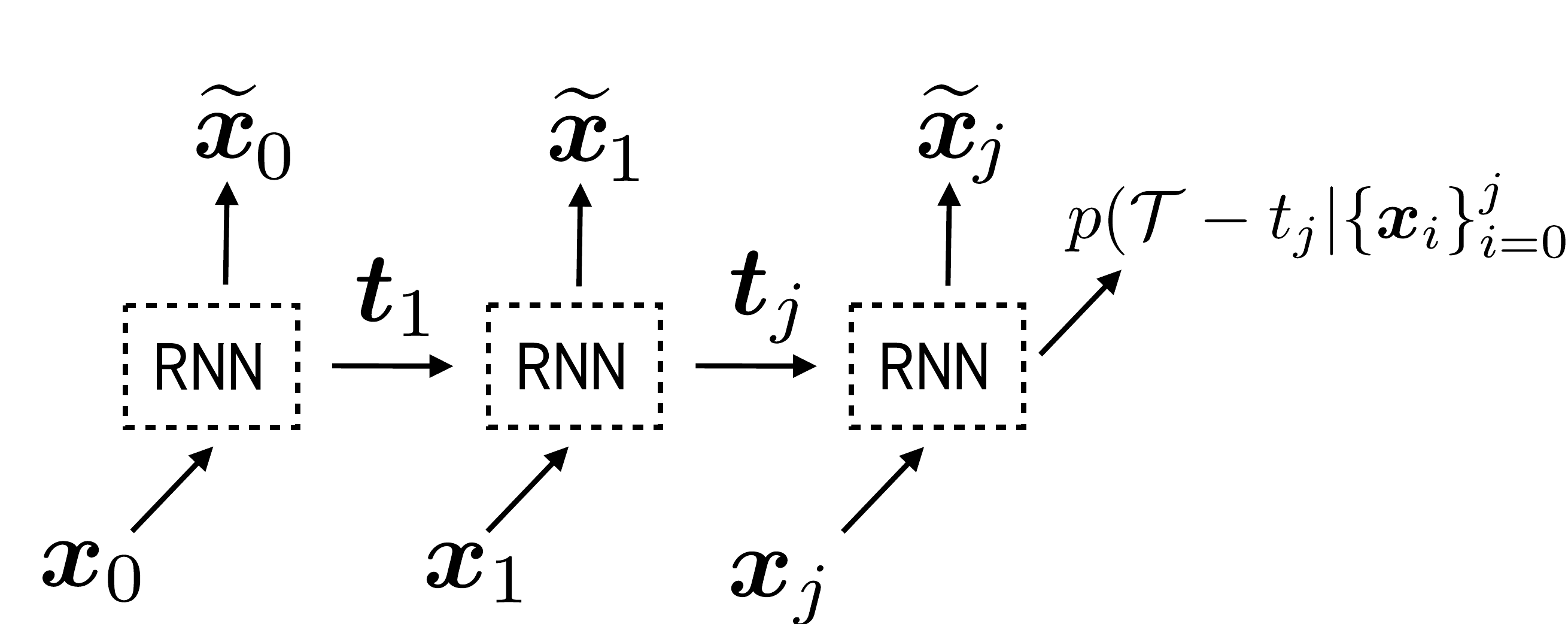}
\captionof{figure}{Time Varying Survival Regression}
\label{fig:tv-surv-reg}
\end{minipage}
\end{minipage}
time-to-event $\mathcal{T}_i$, we observe covariates $\vx^{j}_{i}$ at multiple time points $t^{j}_{i}$. At each time-step, $j$ we estimate the distribution of the remaining time-to-event $T_i - t^{j}_{i}$. The representations of the input covariates $\widetilde{\vx}^{(j+1)}_{i}$ at time-step $(j+1)$ are functions of the covariates, ${\vx}^{(j+1)}_{i}$ and the representation of the  preceeding time-step $\widetilde{\vx}^{j}_{i}$. \package{} automatically handles appropriate padding and batching for sequences of different lengths providing a convenient external API for time-varying survival regression.

\begin{lstlisting}[escapeinside={(*}{*)}]
features, times, events = datasets.load_dataset('PBC', sequential = True)
model = DeepRecurrentSurvivalMachines(k = 3, hidden = [100], typ = 'RNN', layers = 2)
model.fit(features, times, events)
\end{lstlisting}

\newpage

\section{Phenotyping Survival Data}

\begin{figure}[!h]
\centering
\begin{minipage}{0.315\textwidth}
    \centering
    \textbf{\small Unsupervised Phenotyper}\\
    \includegraphics[width=0.8\textwidth, trim={0cm 17cm 26cm 0.8cm},clip]{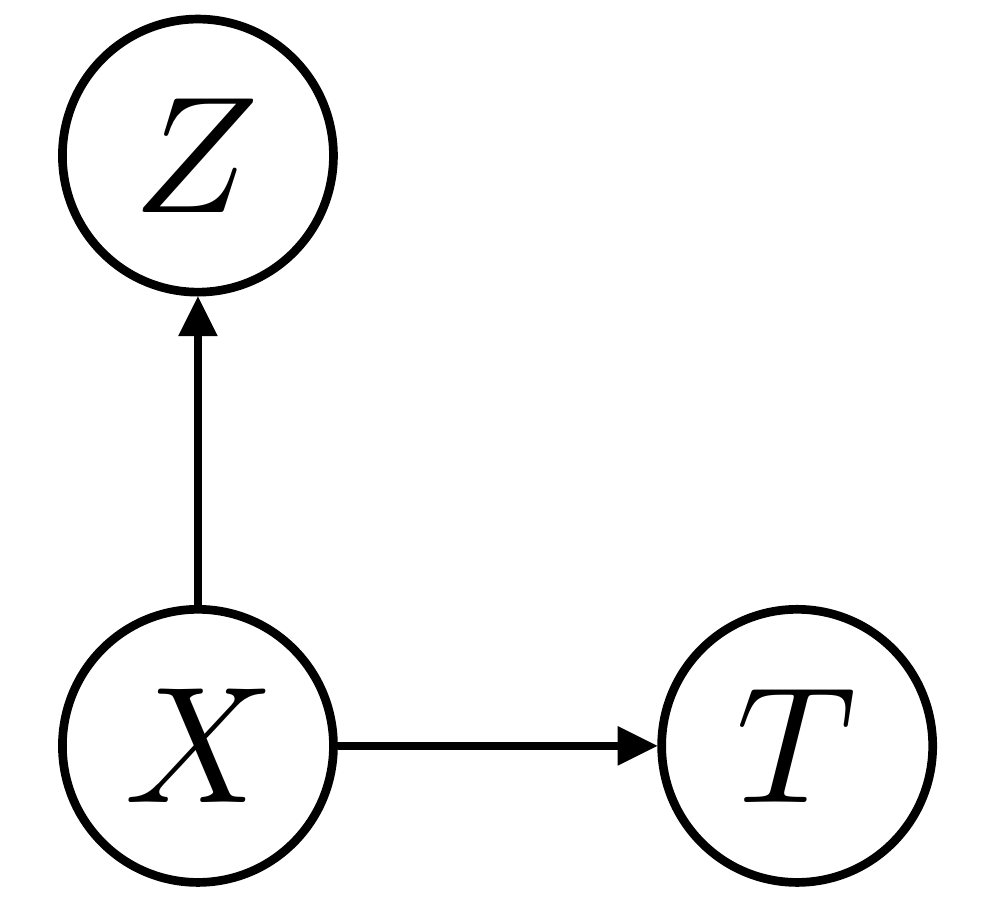}
    \subcaption{Phenotype $Z$ is completely determined by the covariates $X$. The outcome is independent of the phenotype conditioned on the covariates. $(T \perp Z) | X$.}
\end{minipage}
\hfill
\begin{minipage}{0.315\textwidth}
    \centering
    \textbf{\small Supervised Phenotyper}\\
    \includegraphics[width=.8\textwidth, trim={13cm 17cm 13cm 0.8cm},clip]{figures/phenotyping_graph_models.pdf}
    \subcaption{The phenotype $Z$ effects the outcome, hence $ T \not  \perp  Z | X$ . In this case, inference for $Z$ requires knowledge of the distribution of time-to-events $T$.}
\end{minipage}
\hfill
\begin{minipage}{0.315\textwidth}
    \centering
    \textbf{\small Counterfactual Phenotyper}\\
    \includegraphics[width=0.8\textwidth, trim={26cm 17cm 0 0.8cm},clip]{figures/phenotyping_graph_models.pdf}
    \subcaption{The counterfactual phenotype effects the potential outcome under treatment, $T_{(1)}$, but is independent of the outcome under control, $T_{(0)}$.}
\end{minipage}
\caption{{\textbf{DAG representations of the probabilitic phenotypers in \package{}}: $X$ represents the covariates, $T$ the time-to-event and $Z$ is the phenotype to be inferred.}}
    \label{fig:phenotyping-dags}
\end{figure}

Event incidence (survival) rates differ across groups of individuals with heterogeneous characteristics. Identifying groups of patients with similar survival rates can be used to  derive insight about practices and interventions that can help improve longevity for such groups. While domain knowledge can help identify such subgroups, in practice there could be potentially complex, non-linear feature interactions that determine assignment to such groups, making identifiability difficult. In \package{}, we refer to this group identification and survival assessment as \textit{phenotyping}.

\package{} offers multiple approaches to phenotyping that involve either the use of specific domain knowledge as in the case of the intersectional phenotyper, or a completely unsupervised approach that clusters based on the observed covariates. Additionally, \package{} also offers phenotypers that explicitly involve supervision in the form of the observed  outcomes and counterfactual to better inform the learnt phenotypes. \hyperref[fig:phenotyping-dags]{Figure \ref*{fig:phenotyping-dags}} demonstrates the probabilistic phenotypers offered by \package{} as well as the corresponding assumption of conditional independence encoded in the models.\\

\noindent \begin{minipage}{0.5\textwidth}
\subsection{Intersectional Phenotyping}
The \texttt{IntersectionalPhenotyper} class recovers groups, or phenotypes, of individuals over exhaustive combinations of user-specified categorical and numerical features. Numeric covariates are binned based on user-specified quantiles. The intersectional phenotyper is \textit{unsupervised}, but does not have an explicit probabilistic interpretation. \hyperref[fig:intersectional]{Figure \ref*{fig:intersectional}} presents the intersectional phenogroups of cancer status and age group extracted on SUPPORT.
\end{minipage}\hfill
\begin{minipage}{0.48\textwidth}
\centering
\includegraphics[width=.75\textwidth, clip, trim={0cm 0.5cm 1.0cm 0.5cm}]{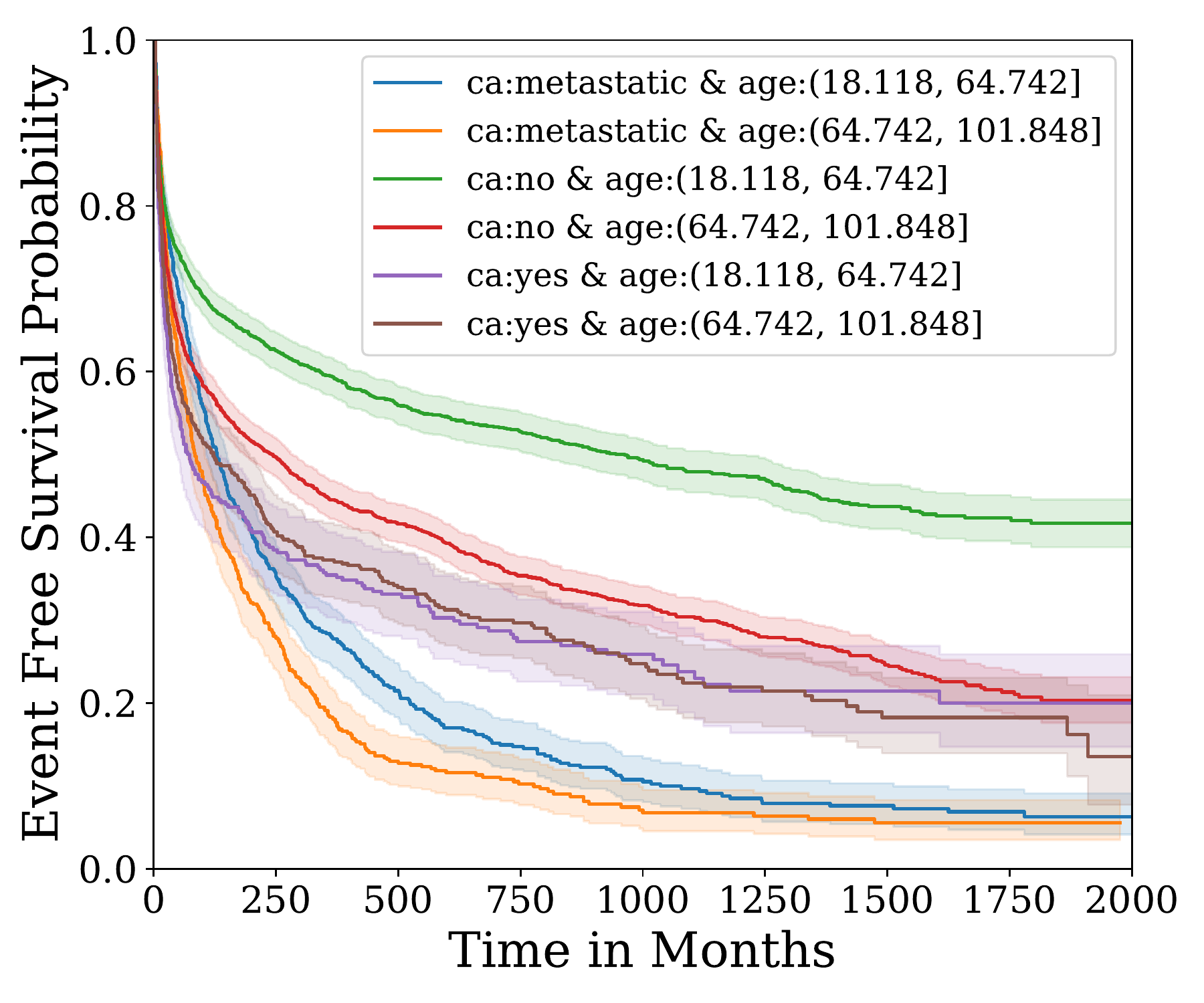}
\captionof{figure}{\small Intersectional Phenotypes.}
\label{fig:intersectional}
\end{minipage}
\newpage
\begin{lstlisting}[escapeinside={(*}{*)}]
from auton_survival.phenotyping import IntersectionalPhenotyper   (*\phantom{A} \hfill \href{https://www.cs.cmu.edu/~chiragn/redirects/auton_survival/phenotyping.html}{{\textbf{\texttt{[Demo Notebook]}}}}*)
# 'ca' is cancer status. 'age' is binned into two quantiles.
phenotyper = IntersectionalPhenotyper(num_vars_quantiles=(0, .5, 1.0), 
                        cat_vars=['ca'], num_vars=['age'])
phenotypes = phenotyper.fit_phenotype(features)
# Plot the phenogroup specific Kaplan-Meier survival estimate.
auton_survival.reporting.plot_kaplanmeier(outcomes, phenotypes)
\end{lstlisting}

\subsection{Unsupervised Phenotyping}

Unsupervised phenotyping  identifies groups based on structured similarity in the space of features. The \texttt{ClusteringPhenotyper} class achieves this by first performing dimensionality reduction of the input covariates $\vx$, followed by clustering. The estimated probability of an individual to belong to a latent group $Z$ is computed as the distance to the cluster normalized by the sum of distance to other clusters. Thus,
\begin{align}
\mathbb{P}(Z=k | X=\vx) = \frac{\mathbf{d}(\vx, \bar{\vx}_k)}{\mathop{\sum}_{j} \mathbf{d}(\vx, \bar{\vx}_j)}.
\end{align}

\noindent Here $\mathbf{d}(\cdot)$ is the distance function corresponding to the underlying clustering method (for instance, `\textit{euclidean}' for $K$-means and `\textit{mahalonobis}' for a Gaussian Mixture Model.)

% Where $d_i$ is the distance to a cluster $k$ for i $\in$ \{1, 2, ..., $n$\} where $i$ $\not=$ $c$.

\begin{lstlisting}[escapeinside={(*}{*)}]
from auton_survival.phenotyping import ClusteringPhenotyper (*\phantom{A} \hfill \href{https://www.cs.cmu.edu/~chiragn/redirects/auton_survival/phenotyping.html}{{\textbf{\texttt{[Demo Notebook]}}}}*)
# Dimensionality reduction using Principal Component Analysis (PCA) to 8 dimensions.
dim_red_method, = 'pca', 8
# We use a Gaussian Mixture Model (GMM) with 3 components and diagonal covariance.
clustering_method, n_clusters = 'gmm', 3 
# Initialize the phenotyper with the above hyperparameters.
phenotyper = ClusteringPhenotyper(clustering_method=clustering_method, 
                                      dim_red_method=dim_red_method, 
                                      n_components=n_components, 
                                      n_clusters=n_clusters)
# Fit and infer the phenogroups.
phenotypes = phenotyper.fit_phenotype(features)
# Plot the phenogroup specific Kaplan-Meier survival estimate.
auton_survival.reporting.plot_kaplanmeier(outcomes, phenotypes)
\end{lstlisting}

\subsection{Supervised Phenotyping}

\begin{figure}[!t]
\begin{minipage}{0.485\textwidth}
    \centering
    \textbf{Unsupervised Phenotyping}
    \includegraphics[width=1\textwidth, clip, trim={0 0 1cm 1.5cm}]{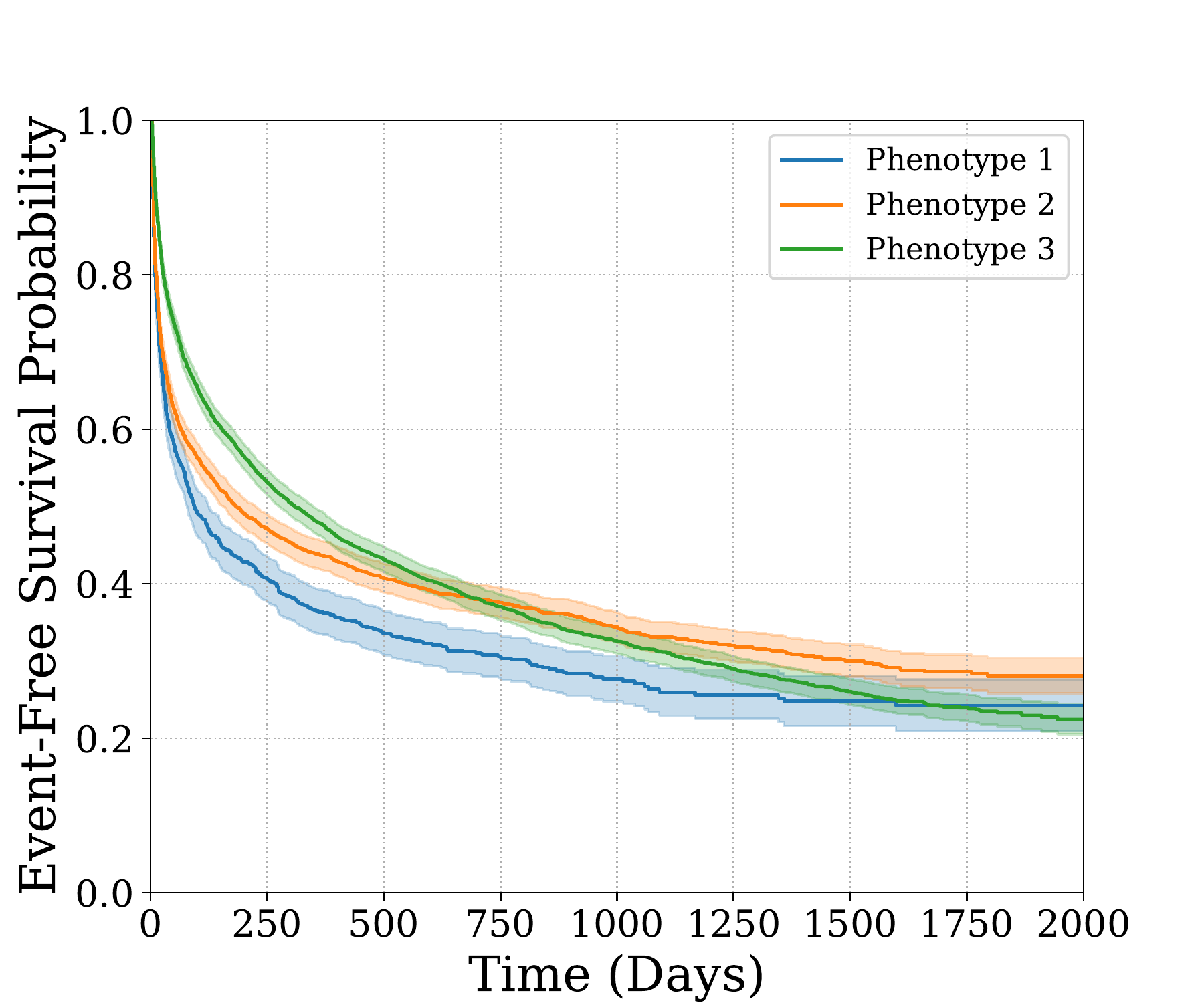}
\end{minipage}
\hfill
\begin{minipage}{0.485\textwidth}
    \centering
   \textbf{Supervised Phenotyping}
    \includegraphics[width=1\textwidth, clip, trim={0 0 1cm 1.5cm}]{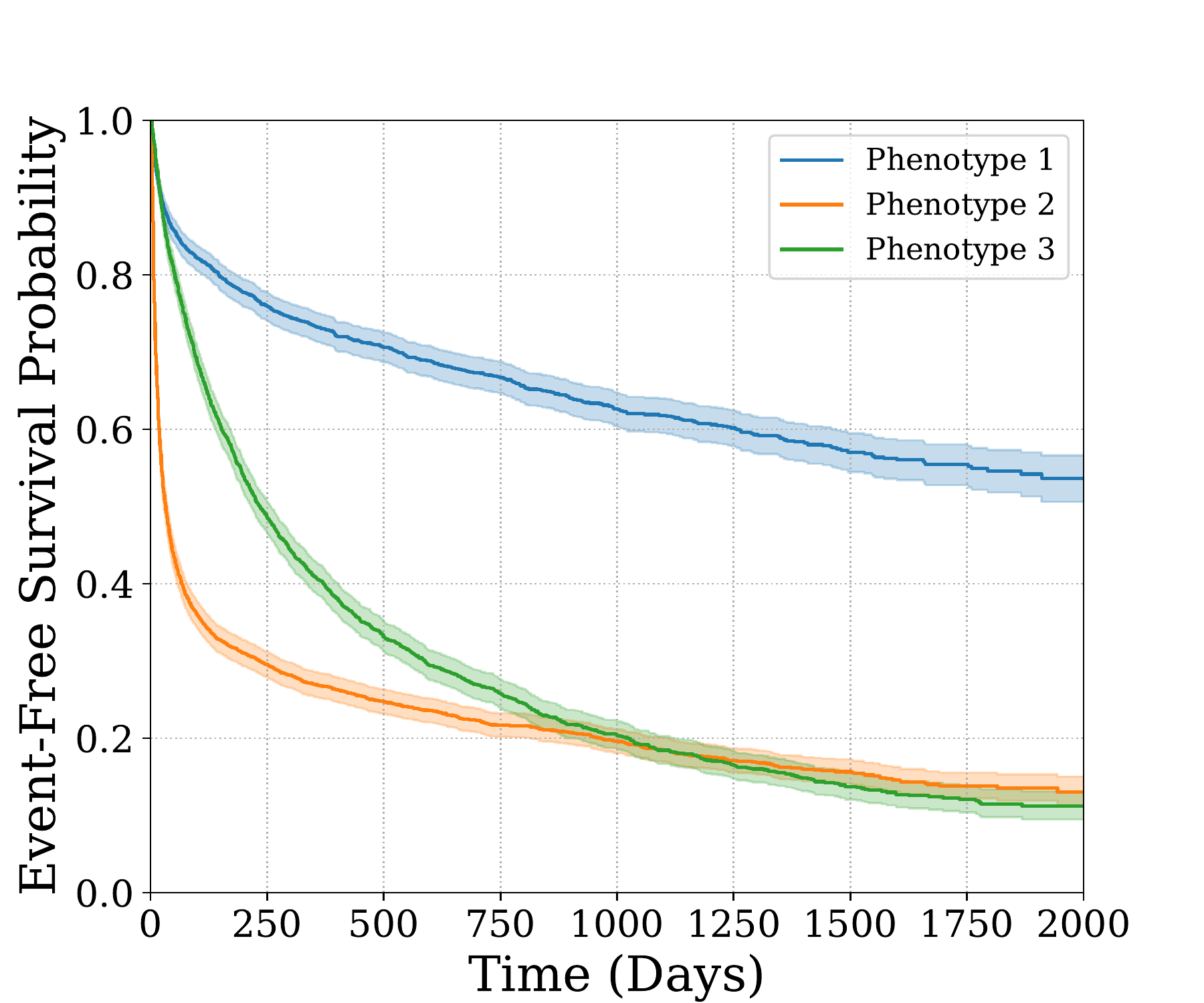}
    \centering
\end{minipage}

\vspace{2mm}
\centering
\textbf{Phenotyping Purity}
\begin{tabular}{r|c|c|c|c}
    \toprule
    \multirow{2}{*}{\textbf{Phenotyper}} & \multicolumn{3}{c|}{\textbf{Brier Score (BS)}} & \textbf{Integrated BS} \\ \cline{2-5}
    {} & \textbf{1 year} & \textbf{2 years} & \textbf{5 years} & \textbf{5 years}\\
    \bottomrule
    {Unsupervised} & 0.246 $\pm$ 0.001 & 0.230 $\pm$ 0.004 & 0.187 $\pm$ 0.015 & 0.218 $\pm$ 0.004 \\
    {Supervised} & 0.234 $\pm$ 0.002 & 0.219 $\pm$ 0.004 & 0.177 $\pm$ 0.014 & 0.207 $\pm$ 0.004\\
    \bottomrule 
\end{tabular}

\caption{Supervised phenotyping with \package{} extracts phenogroups in SUPPORT with higher discriminative power as indicated by the Kaplan-Meier curves and the lower Phenotyping Purity. The intersecting survival curves suggest the phenotypers' ability to recover phenogroups that do not strictly adhere to assumptions of Proportional Hazards.}
\label{fig:latent-phenotyping}
\end{figure}

Supervised phenotyping seeks to explicitly identify latent groups of individuals with similar survival outcomes. Unlike unsupervised clustering, inferring supervised phenotypes requires time-to-events and the corresponding censoring indicators along with the covariates.

\package{} provides utilities to perform supervised phenotyping as a direct consequence of training the \textit{Deep Survival Machines} and \textit{Deep Cox Mixtures} latent variable survival regression estimators. The inferred phenotypes can be obtained after learning these models using the \texttt{predict\_latent\_z} method. Note that these methods differ in the semantic meaning of the phenotypes they recover. DSM recovers phenotypes with similar parametric characteristics while DCM recovers phenotypes that adhere to proportional hazards.

\newpage
\begin{lstlisting}[escapeinside={(*}{*)}]
from auton_survival.models.dcm import DeepCoxMixtures (*\phantom{A} \hfill \href{https://www.cs.cmu.edu/~chiragn/redirects/auton_survival/phenotyping.html}{{\textbf{\texttt{[Demo Notebook]}}}}*)
# Instantiate a DCM Model with 3 phenogroups and a single hidden layer with size 100.
model = DeepCoxMixtures(k = 3, layers = [100])
model.fit(features, outcomes.time, outcomes.event, iters = 100, learning_rate = 1e-4)
# Infer the latent Phenotpyes
latent_z_prob = model.predict_latent_z(features)
# Plot the phenogroup specific Kaplan-Meier survival estimate.
auton_survival.reporting.plot_kaplanmeier(outcomes, latent_z_prob.argmax(axis=1))
\end{lstlisting}

\noindent \textbf{Quantitative Evaluation of Phenotyping}

We measure a phenotyper's ability to extract subgroups, or phenogroups, with differential survival rates by fitting a Kaplan-Meier estimator within each phenogroup followed by estimating the (Integrated) Brier Score within each phenogroup. We refer to this as the \textit{phenotyping purity}. \hyperref[fig:latent-phenotyping]{Figure \ref*{fig:latent-phenotyping}} demonstrates the use of phenotyping purity to compare discriminative power between unsupervised and supervised phenotyping.

\begin{lstlisting}[escapeinside={(*}{*)}]
from auton_survival.metrics import phenotype_purity (*\phantom{A} \hfill \href{https://www.cs.cmu.edu/~chiragn/redirects/auton_survival/phenotyping.html}{{\textbf{\texttt{[Demo Notebook]}}}}*)
# Measure phenotype purity at event horizons of 1, 2 and 5 years. 
phenotype_purity(phenotypes, outcomes, strategy='instantaneous', time=[365,730,1825])
# Measure phenotype purity at an event horizon of 5 years. 
phenotype_purity(phenotypes, outcomes, strategy='integrated', time=1825)
\end{lstlisting}

% \subsection{\href{https://nbviewer.org/github/autonlab/auton-survival/blob/master/examples/Demo\%20of\%20CMHE\%20on\%20Synthetic\%20Data.ipynb}{{\textbf{\texttt{Demo Notebook}}}}}
\newpage
\subsection{Counterfactual Phenotyping}

 In real world clinical scenarios, individuals demonstrate heterogeneous treatment effects to an intervention. Thus, clinical decision support often requires reasoning about which groups of individuals benefit \textit{least} (or \textit{most}) from a certain intervention. Such identification is often referred to as \textit{subgroup discovery} or \textit{subgroup analysis}. While there has been extensive study in subgroup recovery with binary or continuous outcomes \citep{dusseldorp2014qualitative, lipkovich2011subgroup, foster2011subgroup, nagpal2020interpretable, wang2022causal}, censored time-to-event outcomes are less studied in the context of phenotyping.
 
 More formally we can describe counterfactual phenotyping as estimating a function $f(\vx): \mathbb{R}^d \to \{0, 1\}$
\begin{align}
    f = \mathop{\textrm{arg min}}_{f\in \mathcal{F}} \mathbb{CATE}( \mathcal{D} | f(X) = 1 ) \quad \textrm{such that} \quad \mathbb{E}[f(X)] > \alpha .
\end{align}
The first term here refers to the (Conditional) Average Treatment Effect within the discovered phenotype, while the second term controls the size of the phenotype.\\

\noindent \textbf{Virtual Twins Survival  Regression:} \package{} includes a Virtual Twins model \citep{foster2011subgroup} involving first estimating the potential outcomes under treatment and control using a counterfactual Deep Cox Proportional Hazards model, followed by regressing the difference in the estimated counterfactual Restricted Mean Survival Times using a Random Forest regressor.

\begin{lstlisting}[escapeinside={(*}{*)}]
from auton_survival.phenotyping import SurvivalVirtualTwins 
# Instantiate the Survival Virtual Twins
model = SurvivalVirtualTwins(horizon=365)
model.fit(features, outcomes.time, outcomes.event, interventions)
# Infer the estimated counterfactual phenotype probability.
phi_probs = model.predict_latent_phi(features)
\end{lstlisting}
\vspace{0.1em}

\noindent \textbf{Cox Mixture Latent Variable Model:} \package{} also includes an implementation of the Cox Mixtures with Heterogenous Effects (CMHE) \citep{nagpal2022counterfactual} model for counterfactual phenotyping. The CMHE model involves modelling the individual hazard rate under an intervention $(A=\va)$ as
\begin{align}
\underbrace{\bm{\lambda}(t|X=\vx, {\color{black} Z=k},{\color{black} \bm{\phi}=m},  A=\bm{a})}_{\textrm{Conditional Hazard Rate}} = \overbrace{\bm{\lambda}_k(t)}^{\mathclap{\text{\shortstack{Base Survival Rate}}}} {\color{black}\underbrace{\exp\big(h^{k}_{\bm\theta}(\vx)\big)}_{\mathclap{\text{\shortstack{Effect of Confounders}}} }}{\color{black} \overbrace{\exp(\omega_m)^{\color{black}\bm{a}}}^{\mathclap{\text{\shortstack{Treatment Effect}}}}},
\end{align}

where $(Z=k)$ is the base survival rate,  $(\bm{\phi}=m)$ is the latent subgroup memberships that mediates the treatment effect. 

\begin{lstlisting}[escapeinside={(*}{*)}]
from auton_survival.models import cmhe (*\phantom{A} \hfill \href{https://www.cs.cmu.edu/~chiragn/redirects/auton_survival/cmhe_notebook.html}{{\textbf{\texttt{[Demo Notebook]}}}}*)
# Instantiate the Cox Mixture with Heterogenous Effects model
model = cmhe.DeepCoxMixturesHeterogenousEffects(k=1, g=2, layers=[100])
model.fit(features, outcomes.time, outcomes.event, interventions, 
           iters=iters, learning_rate=learning_rate, batch_size=batch_size)
# Infer the estimated counterfactual phenotypes.
latent_phi_probs = model.predict_latent_phi(features)
\end{lstlisting}
\vspace{0.1em}

% \begin{minipage}{1\textwidth}
% aEDAWSEFAW
% \end{minipage}
% \begin{minipage}{1\textwidth}
% aEDAWSEFAW
% \end{minipage}

% \subsubsection{\href{https://nbviewer.org/github/autonlab/auton-survival/blob/master/examples/Demo\%20of\%20CMHE\%20on\%20Synthetic\%20Data.ipynb}{{\textbf{\texttt{Demo Notebook}}}}}

\newpage
\section{Evaluation}
\subsection{Censoring-Adjusted Evaluation Metrics}

The survival model performance can be evaluated using the following metrics, among others. \package{} offers utilities to compute these metrics along with bootstrapped confidence intervals with a convenient API.\\

\noindent \textbf{Brier Score} (BS): {The Brier Score involves computing the Mean Squared Error of the estimated survival {probabilities, $\widehat{\mathbb{P}}(T>t|X=\vx)= f(\vx, t)$ at a specified time horizon, $t$.} As a proper scoring rule, Brier Score gives a sense of both discrimination and calibration.} 

{Under the assumption that the censoring distribution is independent of the time-to-event, we can obtain an unbiased, censoring adjusted estimate of the Brier Score $\widehat{\text{BS}}_{\text{IPCW}}(t)$ using inverse probability of censoring weights (IPCW) from a Kaplan-Meier estimator of the censoring distribution, $\hat G(.)$ as proposed in \cite{graf1999assessment, gerds2006consistent}.}
\begin{align*}
\text{BS}(t) &= \mathbb{E}_{\mathcal{D}}\big[ \big(\bm{1}\{ T_i > t \} - \widehat{\mathbb{P}}(T>t|X=\vx)\big)^2  \big]\\
\widehat{\text{BS}}_{\text{IPCW}}(t) &= \frac{1}{n}\sum_{i=1}^{n}\bigg[ \frac{f(\vx_i, t)^2 \bm{1}\{ T \leq t, \delta_i =1 \}}{\hat G_i(T_i)} +  \frac{ \big(1-f(\vx_i, t))^2 \bm{1}\{ T > t \}}{\hat G_i(t)} \bigg].
\end{align*}

\noindent \textbf{Area under ROC Curve} (AUC): {ROC Curves are extensively used in classification tasks where the true positive rate (TPR) is plotted against the false positive rate (1-true negative rate, or TNR) to measure the in model discriminative power over all output thresholds.}

{To enable the use of ROC curves to assess the performance of survival models subject to censoring, we employ the technique proposed by \cite{uno2007evaluating, hung2010optimal}, which treats the TPR as time-dependent on a specified horizon, $t$ and adjusts survival probabilities, $f(\vx, t)$, using the IPCW from a Kaplan-Meier estimator of the censoring distribution, $\hat G(t)$. Estimating the TNR requires observing outcomes for each individual, only uncensored instances are used. Refer to \cite{kamarudin2017time} for details on computing ROC curves in the presence of censoring.}
$$\widehat{\text{TPR}}(c, t)= \frac{\sum\limits_{i=1}^{n}  \frac{\delta_i}{\hat{G}(T_i)} \cdot \bm{1}\{ f(\vx_i, t) > c, T_i \leq t  \}    }{\sum\limits_{i=1}^{n} \frac{\delta_i}{\hat G(T_i)}\cdot \bm{1}\{T_i < t  \} \cdot};\quad \widehat{\text{TNR}}(c, t)=\frac{\sum\limits_{i=1}^{n}  \bm{1}\{ f(\vx_i, t) \leq c, T_i > t  \}    }{\sum\limits_{i=1}^{n} \bm{1}\{T_i > t  \}}$$

\noindent\textbf{Time Dependent Concordance Index} ($C^\text{td}$): 
Concordance Index compares risks across all pairs of individuals within a fixed time horizon, $t$, to estimate ability to appropriately rank instances relative to each other in terms of their risks, $f(\vx, t)$. 
\begin{align*}
C^{td }(t) = \mathbb{P}\big( f(\vx_i, t) \leq f(\vx_j, t) | \delta_i=1, T_i<T_j, T_i \leq t \big) 
\end{align*}

We employ the censoring adjusted estimator for $C^{\text{td}}$ that exploits IPCW estimates from a Kaplan-Meier estimate of the censoring distribution. Further details can be found in \cite{uno2011c} and \cite{gerds2013estimating}.
\newpage

% \noindent \textbf{Expected $\mathbf{\ell_1}$ Calibration Error }(ECE): The ECE measures the average absolute difference between the observed and expected (according to the risk score) event rates, conditional on the estimated risk score. At time $t$, let the predicted risk score be $R(t) = \widehat{\mathbb{P}}(T>t | X)$. Then, the ECE approximates
%  \begin{align*}
% \text{ECE}(t) = \mathbb{E} \big[ \big| \mathbb{P}(T > t | R(t)) - R(t)  \big| \big]
% \end{align*}
% by partitioning the risk scores $R$ into $q$ quantiles $\{[r_j, r_{j+1})\}_{j=1}^q$. and computing the Kaplan-Meier estimate of the event rate $\text{KM}_j(t) \approx P(T > t | R \in [r_j, r_{j+1}))$, and the average risk score $\overline{R}_j = \frac{q}{n} \sum_{i : R_i \in [r_j, r_{j+1})} R_i$ in each bin. Altogether, the estimated ECE is
% \begin{equation*}
%     \widehat{\text{ECE}}(t) = \frac{1}{q} \sum_{j = 1}^q |\text{KM}_j(t) - \overline{R}_j(t)|.
% \end{equation*}
% In practice, we fix the number of quantiles to be 20 for our experiments.\\

\begin{lstlisting}
from auton_survival.metrics import survival_regression_metric
# Infer event-free survival probability from model
predictions = model.predict_survival(features, times)
# Compute Brier Score, Integrated Brier Score
# Area Under ROC Curve and Time Dependent Concordance Index
metrics = ['brs', 'ibs', 'auc', 'ctd']
for metric in metrics
    score = survival_regression_metric(metric, outcomes_train, outcomes_test,                                           predictions_test, times=times)
\end{lstlisting}

% \subsection{Discrimination?}

% \subsection{Calibration?}

\subsection{Comparing Treatment Arms} 
\label{sec:3.2}

\begin{figure}[!h]
\begin{minipage}{0.315\textwidth}
    \includegraphics[width=1\textwidth]{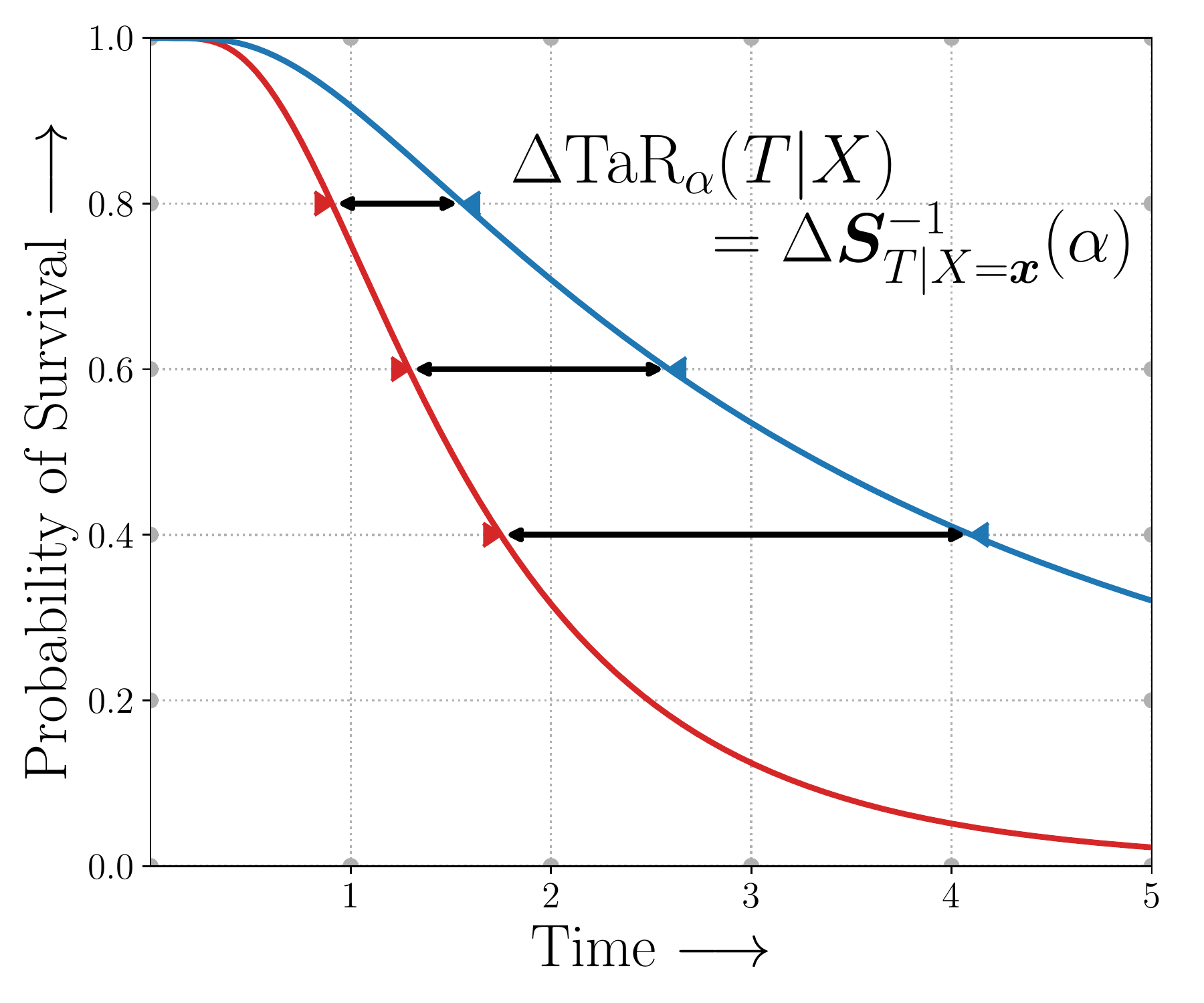}%
    \subcaption{Time at Risk ($\textrm{TaR}$)}
    \label{fig:TaR}
\end{minipage}
\hspace{1mm}
\begin{minipage}{0.315\textwidth}
    \includegraphics[width=1\textwidth]{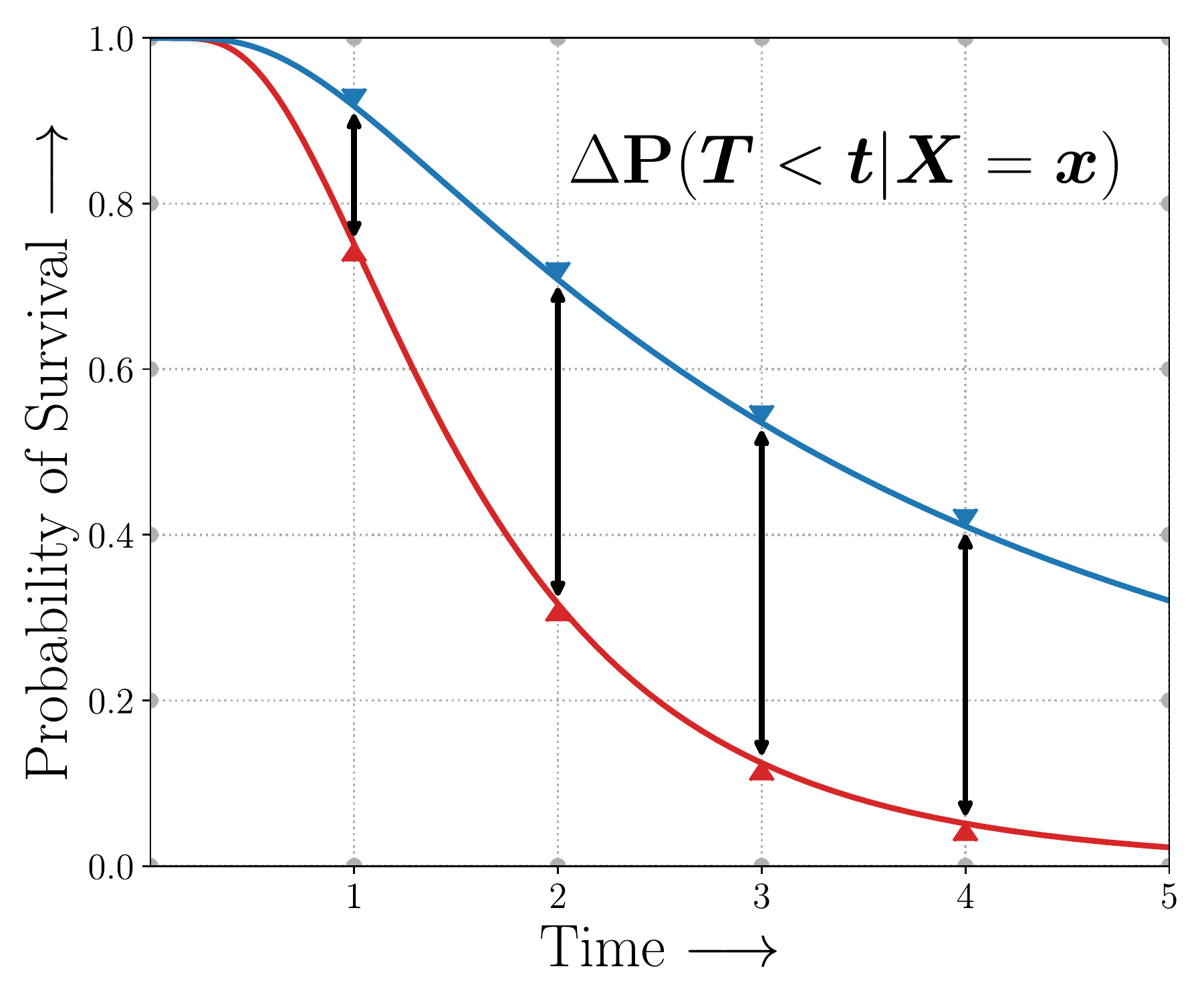}%
    \subcaption{Risk at Time}
    \label{fig:riskattime}
\end{minipage}
\hspace{1mm}
\begin{minipage}{0.315\textwidth}
    \includegraphics[width=1\textwidth]{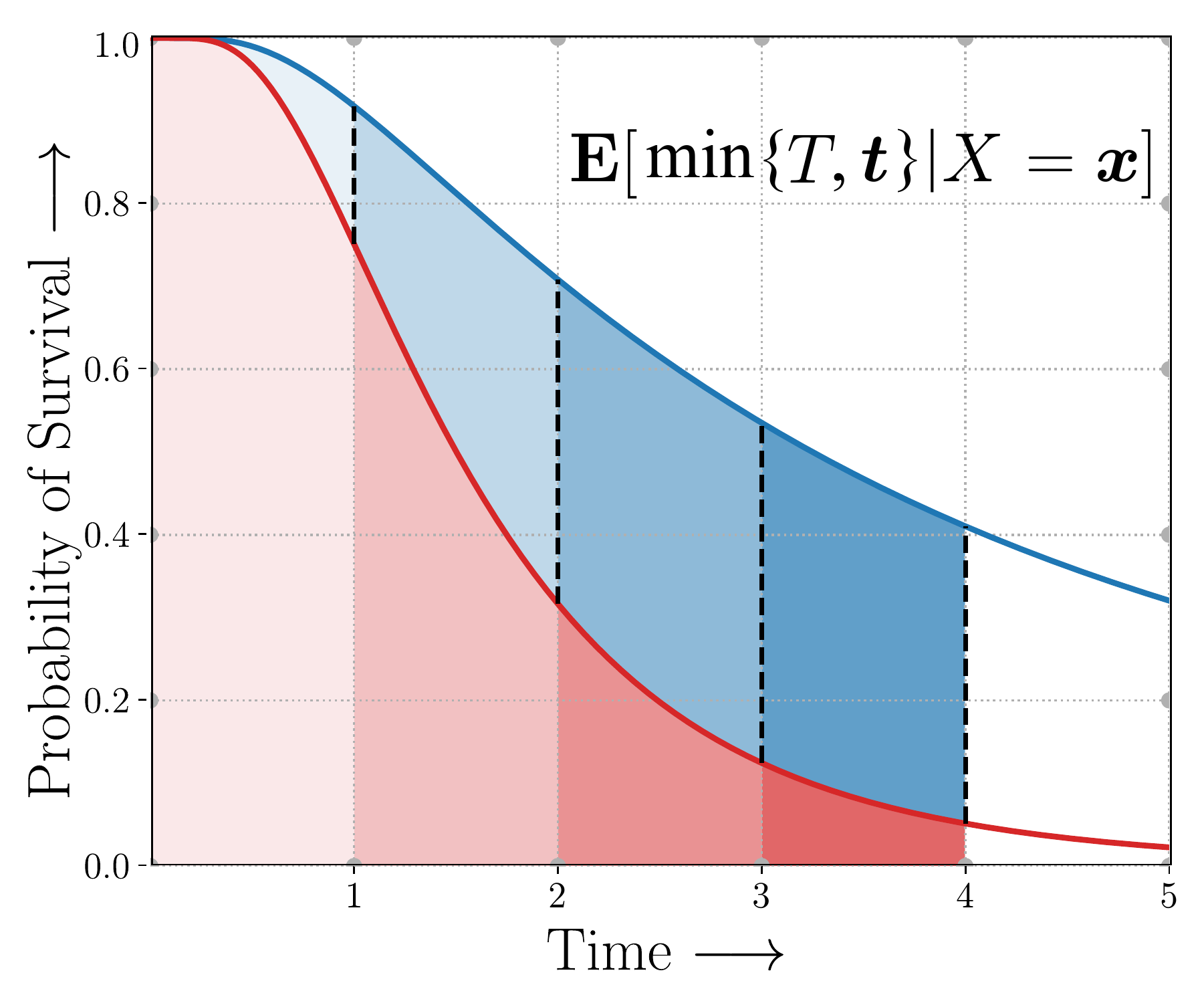}%
    \subcaption{{RMST}}
    \label{fig:RMST}
\end{minipage}
\caption{Treatment effects measured in terms of the difference in metrics computed for treatment and control groups including (a) the time at a specified level of risk (b) risk at a certain time and (c) the expected survival time over a truncated time horizon.}
\end{figure}

\noindent{\textbf{Hazard Ratio}}

Assuming the proportional hazards assumptions holds, the treatment effect can be modelled as the ratio of hazard rates between the treatment and control arms. This corresponds to fitting a univariate Cox Proportional Hazards model and is arguably the most popular approach in estimating treatment effects for censored time-to-events.\\ 
% \begin{align}
%     \textrm{Hazard-Ratio} = \frac{\bm{\lambda}(t|\bm{A}=1)}{\bm{\lambda}(t|\bm{A}=0)}
% \end{align}

\noindent{\textbf{Time at Risk (TaR)}}

Time at Risk (TaR) (\hyperref[fig:TaR]{Figure \ref*{fig:TaR}}) measures time-to-event at a specified level of risk. Computing TaR may be of interest when maximum permissible outcome risk is predefined and the corresponding time-to-event is the deciding factor for assessing treatment benefit at that risk level.
\begin{align}
    \operatorname{TaR}_\alpha(T|X)=\sup_t\big\{t\in\mathbb{R}^{+}:\bm{S}_{T|X}(t)<\alpha\big\} = \bm{S}^{-1}_{T|X}(\alpha)
\end{align}
\noindent{\textbf{Risk at Time}} 

Risk at Time (\hyperref[fig:riskattime]{Figure \ref*{fig:riskattime}}) measures risk at a specified time horizon. This metric may be of interest when treating survival as a binary outcome over a specific time horizon.
\begin{align}
   \textrm{Risk-at-Time}(t|X=\vx) = \mathbb{P}(T<t|X=\vx) = 1-\mS(t|X=\vx)
\end{align}

\noindent{\textbf{Restricted Mean Survival Time (RMST)}} 

{The RMST (\hyperref[fig:RMST]{Figure \ref*{fig:RMST}}) is the expected (or mean) time-to-event conditioned on a specified time horizon. Mathematically, the RMST is $\mathbb{E}[\mathrm{min}\{T, \vt\}|X=\vx]$ and translates into the area under the survival curve till time $\vt$.}
\begin{align}
\mathrm{RMST}_{\vt}(T|X=\vx) &= \mathbb{E}[\mathrm{min}\{T, \vt\}|X=\vx]\\
&=\int_0^{t} \mS(\bm{t}| X=\vx)\textrm{d}\bm{t}. \nonumber
\end{align}

RMST might be preferred over Hazard Ratio in comparing treatment arms where there is reason to believe that the proportional hazards assumption is violated.

\begin{lstlisting}
from auton_survival.metrics import treatment_effect
# Compute the difference in RMST between treatment and control groups
effect = treatment_effect(metric='restricted_mean', outcomes=outcomes
                             treatment_indicator=treatment_indicator,
                             weights=None, horizon=120, n_bootstrap=500)
\end{lstlisting}

\subsection{Propensity Adjusted Treatment Effects}

Differences in the values of metrics specified in \hyperref[sec:3.2]{Section \ref*{sec:3.2}} across treatment arms is often averaged over the population to estimate net benefit or the average treatment $(\mathbb{ATE})$,
$$ \mathbb{ATE}(\mathcal{D}, f) = \mathop{\mathbb{E}}_{\vx \sim \mathcal{D}} \big[\mathbb{E}[ f_1(x) - f_0(x) | X = \vx ]\big]  $$
Directly computing the metrics in above in an observational settings would result in mis-estimations of treatment effects as it would not adjust for potential confounders that influence both treatment assignment and the outcome. \package{} supports the computation of propensity-adjusted treatment effects in terms of the above metrics through bootstrap resampling the dataset, $\mathcal{D}$ with replacement. Here the resampling weights are obtained using inverse propensity of treatment weighting (IPTW). The bootstrapped treatment effect thus converges to the Thompson-Horvitz estimate of the Treatment Effect.
\begin{align}
% := \{ (\vx_1,\vt_1,\delta_1)^{*}, \ldots ,(\vx_n,\vt_n,\delta_n)^{*} \}
 \mathbb{ATE}(\mathcal{D}^{*}, f) = \mathop{\mathbb{E}}_{\vx \sim \mathcal{D}^{*}} \big[\mathbb{E}[ f_1(x) - f_0(x) | X = \vx ]\big]; \quad    \mathcal{D}^{*}  \sim \frac{1}{\widehat{\mathbb{P}}(A|X)}\cdot\mathbb{P}^{*}(\mathcal{D})  
 %; \quad \{  (\vx_i,\vt_i,\delta_i)^{*} \}^{n}_{i=1} 
 \end{align}

The \package{} function \texttt{treatment\_effect} function supports sample weight inputs in the form of treatment propensity scores, such as obtained from scaled classification model scores that estimate probability of treatment.

% \begin{figure}
%     \centering
%     \includegraphics{MLHC/plot.png}
%     \caption{Caption}
%     \label{fig:my_label}
% \end{figure}

\begin{lstlisting}
from auton_survival.metrics import treatment_effect
from sklearn.linear_model import LogisticRegression
# Train a classification model to compute "treatment" propensity scores
model = LogisticRegression(penalty='l2').fit(features, treatment_indicator)
treatment_propensity = model.predict_proba(features)[:, 1]
# Compute the treatment effect after adjusting for treatment propensity
adjusted_effect = treatment_effect(metric='hazard_ratio', outcomes=outcomes
                                       treatment_indicator=treatment_indicator,
                                       weights=treatment_propensity, n_bootstrap=500)
\end{lstlisting}

\newpage

\section{Case Study: Regional disparities in Breast Cancer Incidence}

\begin{figure}[h!]
\begin{minipage}{1\textwidth}

    \begin{minipage}{0.5\textwidth}
    \centering
    \textbf{Sample Size and RMST by Region}
    \begin{tabular}{r|c|c} \toprule
        \textbf{Region}&  \textbf{Size} & \textbf{RMST} \\ \midrule
         Greater California & 47,757 & 108.9 $\pm$ 0.261 \\
          New Jersey & 27,665 & 107.3 $\pm$ 0.352\\
         Greater Georgia & 14,212 & 106.8 $\pm$ 0.559\\
         Louisiana & 11,837 & 104.1 $\pm$ 0.598\\ 
         Kentucky & 11,581 & 106.4 $\pm$ 0.610\\ \midrule
         All Regions & 113,052 & 107.5 $\pm$ 0.189 \\ \bottomrule 
    \end{tabular}
    \vspace{6mm}
    % \subcaption{\textcolor{red}{Sampled population size varies across regions and can bias analysis of treatment, or region, effects. (\textbf{RMST: 10 Year Restricted Mean Survival Time.}})}
    \end{minipage}%
    \hfill
    \begin{minipage}{0.45\textwidth}
    \includegraphics[width=1.\textwidth]{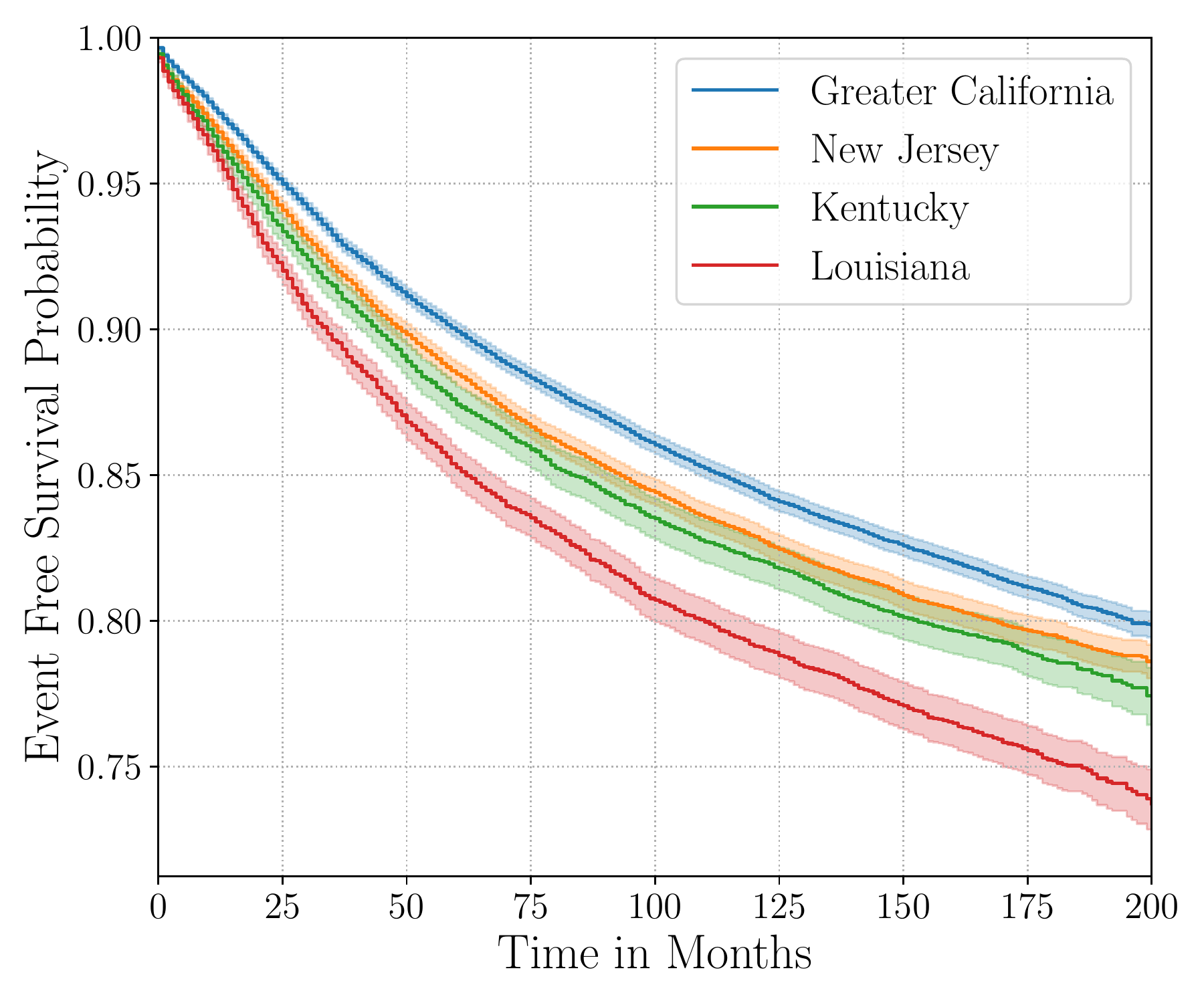}
    % \subcaption{\textcolor{red}{Kaplan Meier estimates show different survival rates across regions.}}
    \end{minipage}%
\end{minipage}
\caption{Breast cancer survival from the cohort subset of the SEER study. Notice discrepancies in the Kaplan-Meier estimated survival rates when stratified by the geographic region. Similarly, regional discrepancies in expected time-to-event are highlighed by mean (95\% CI) RMST computed for bootstrap samples conditioned on a 10-year horizon.}
\label{fig:seer-bc-all}
\end{figure}

The SEER registry established by the National Cancer Institute consists of patient demographic and tumour morphology data and long term incidence information of approximately one-third of the U.S. population \citep{ries2008seer}. This study is a retrospective analysis of the SEER\footnote{Surveillance, Epidemiology and End Results - National Cancer Institute} data. For this study, we consider a subset of 113,052 patients from Greater California, Kentucky, Louisiana, New Jersey and Georgia diagnosed with breast cancer between a four year period from January 2000 to December 2003\footnote{We restrict our study to this cohort subset as it had a consistent coding pattern vis-a-vis tumour morhology and therapy characteristics in the original SEER database}. We consider all-cause mortality as the outcome of interest with a maximum follow up period of 18 years from January, 2000. Along with outcomes we also consider additional variables including age, race, tumour morphology and treatments administered. A complete list of the variables considered in this study can be found in \noindent \hyperref[apx:seer-features]{Appendix \ref*{apx:seer-features}}.

% \subsection{M stratified by Region}

\subsection{The Effect of Geographical Region on Breast Cancer Mortality}

It is evident that there are discrepancies between the observed mortality rates stratified by the geographic region (\hyperref[fig:seer-bc-all]{Figure \ref*{fig:seer-bc-all}}). A natural question arises as to whether this discrepancy can be attributed to geographic region or other socio-economic or physiological confounding factors that effect both belonging to these regions and the outcome. 

We attempt to provide insight into this question by assessing the effect size via the Potential Outcomes framework by considering the region as an intervention. We first estimate the counterfactual survival rates across treatment arms to evaluate the effect of treatment on event-free survival rates. We further verify our findings by comparing treatment effects before and after adjusting for treatment propensity by inverse propensity weighting.

\subsubsection*{Counterfactual Survival Estimation}

\begin{minipage}{0.5\textwidth}

To adjust estimates of survival with counterfactual estimation, we train two separate Deep Cox models as described in \hyperref[sec:counterfactual]{Section \ref*{sec:counterfactual}} on data from Greater California and Louisiana as counterfactual regressors. The fitted regressors are then applied to estimate the survival curves for each instance, which are then averaged over treatment groups to compute the domain-specific survival rate.

\hyperref[fig:seer-bc-cf]{Figure \ref*{fig:seer-bc-cf}} presents the counterfactual survival rates compared with the survival rates obtained from a Kaplan-Meier estimator. The Kaplan-Meier estimator does not adjust for confounding and so overestimates treatment effect as evidenced from the extent that survival rates differ between regions. Counterfactual regression adjusts for confounding factors and predicts more similar survival rates between regions.

\end{minipage}
\hfill
\begin{minipage}{0.45\textwidth}
\centering
    \includegraphics[width=1\textwidth]{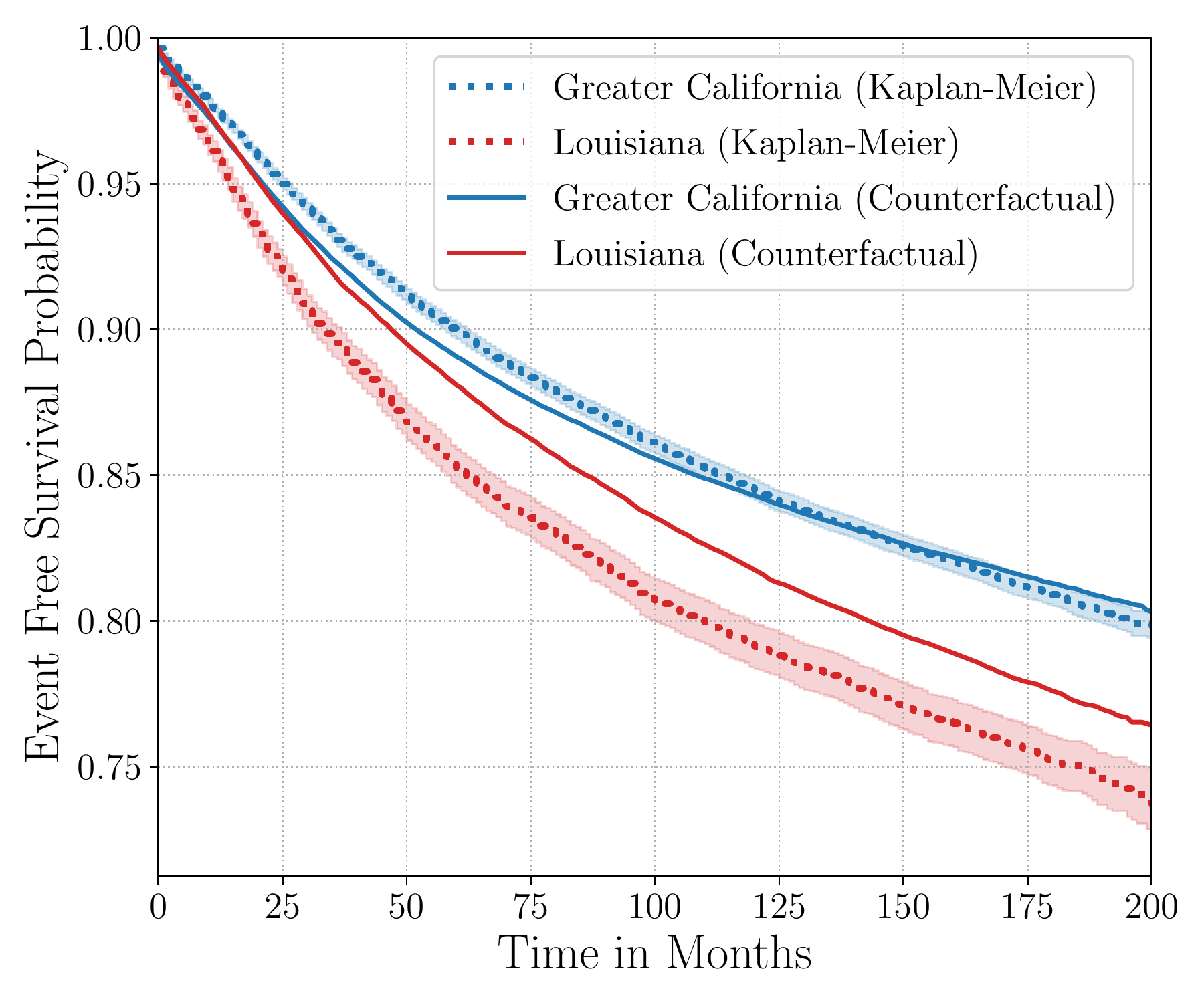}
    \captionof{figure}{{The Kaplan-Meier estimator overestimates effect of region  on mortality compared with the counterfactual regression model, which reduces discrepancy in survival rates between regions.}}
    \label{fig:seer-bc-cf}
\end{minipage}
\vspace{3mm}

%\newpage

\subsubsection*{Propensity Adjusted Treatment Effects}

\begin{figure}[h!]
    \centering
    \begin{minipage}{0.33\textwidth}
    \centering
    \includegraphics[width=1\textwidth]{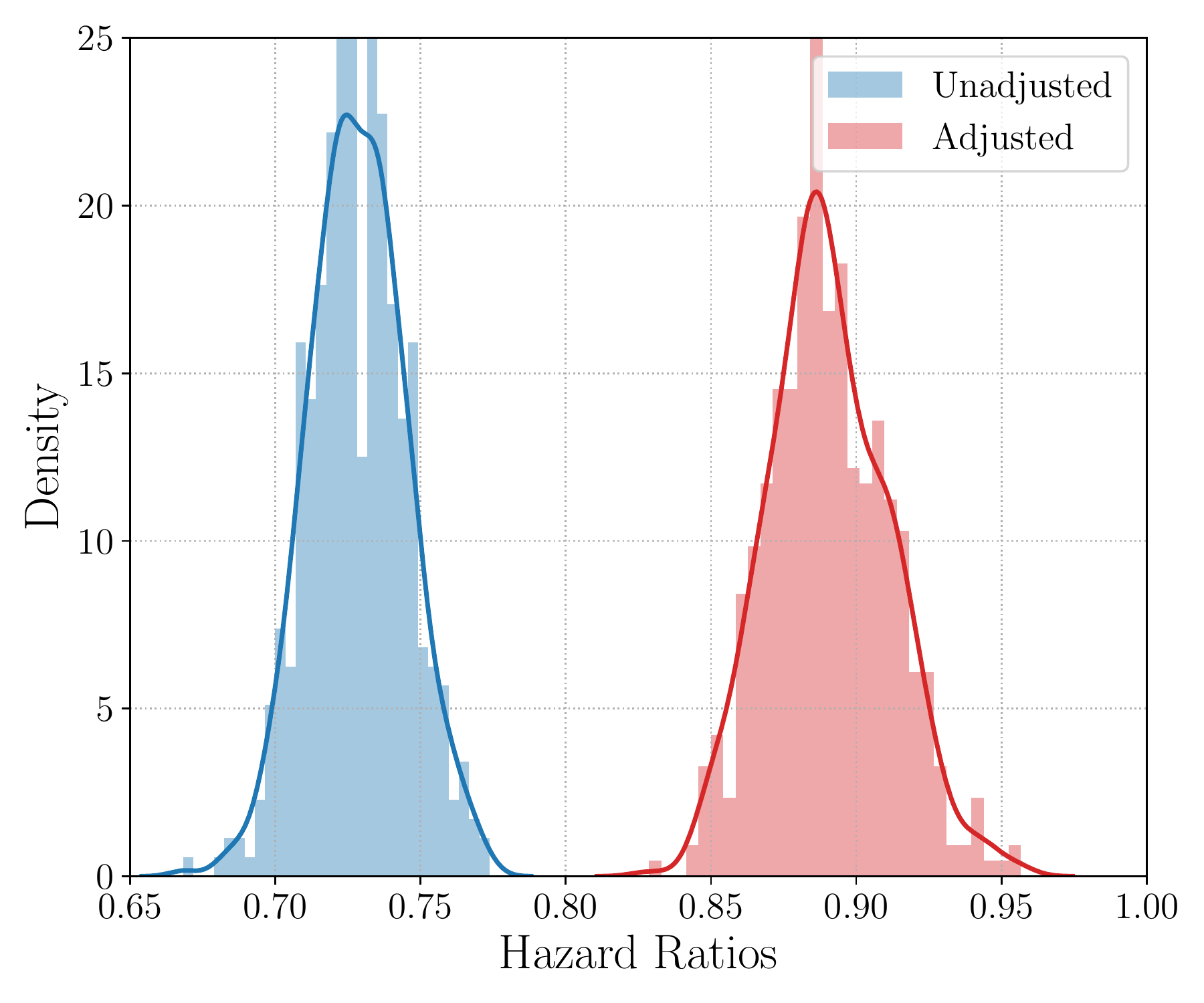}
    \subcaption{Hazard Ratios}
    \end{minipage}\hfill
    \begin{minipage}{0.33\textwidth}
    \includegraphics[width=1\textwidth]{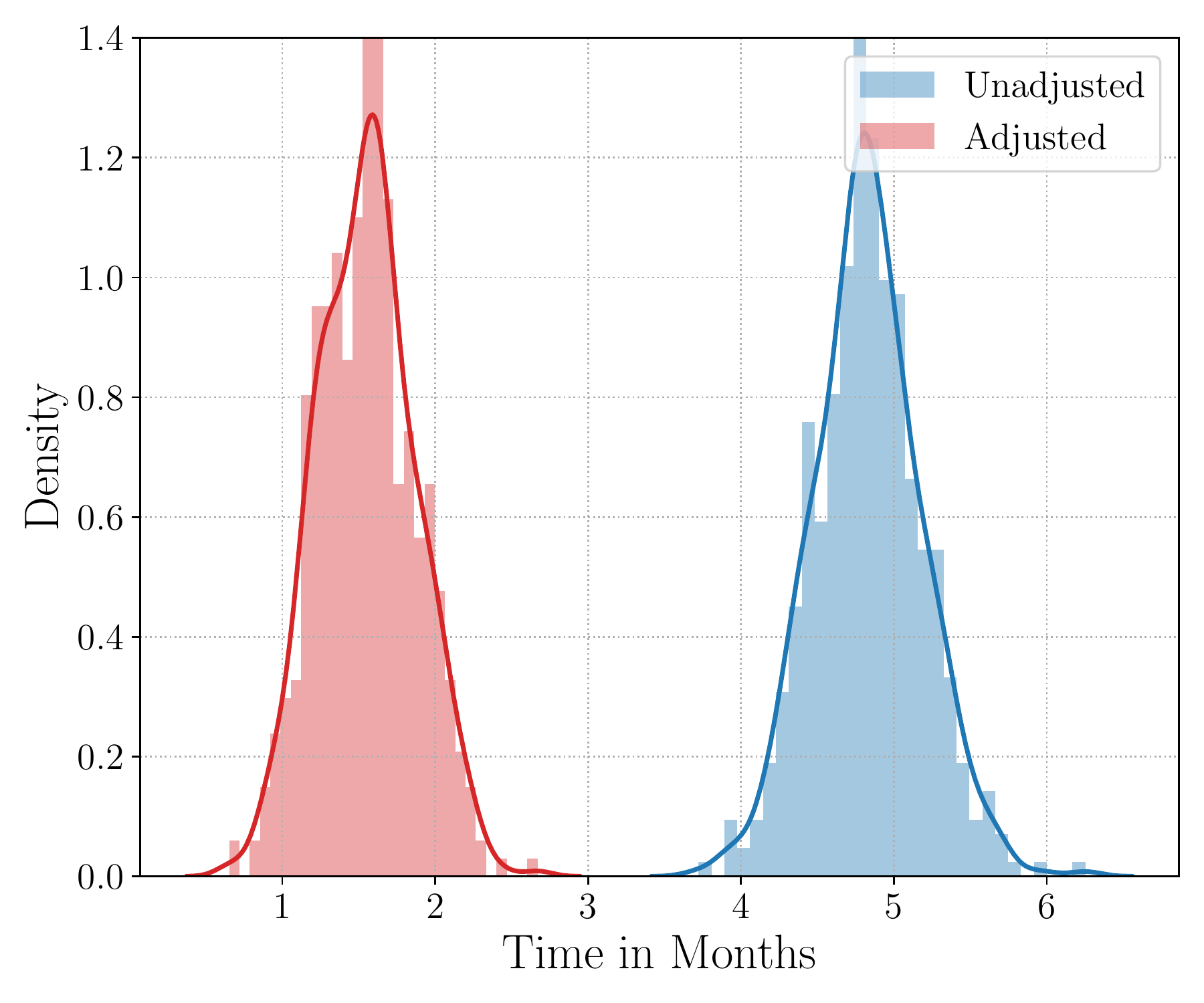}
    \subcaption{RMST}
    \end{minipage}\hfill
    \begin{minipage}{0.33\textwidth}
    \centering
    \includegraphics[width=1\textwidth]{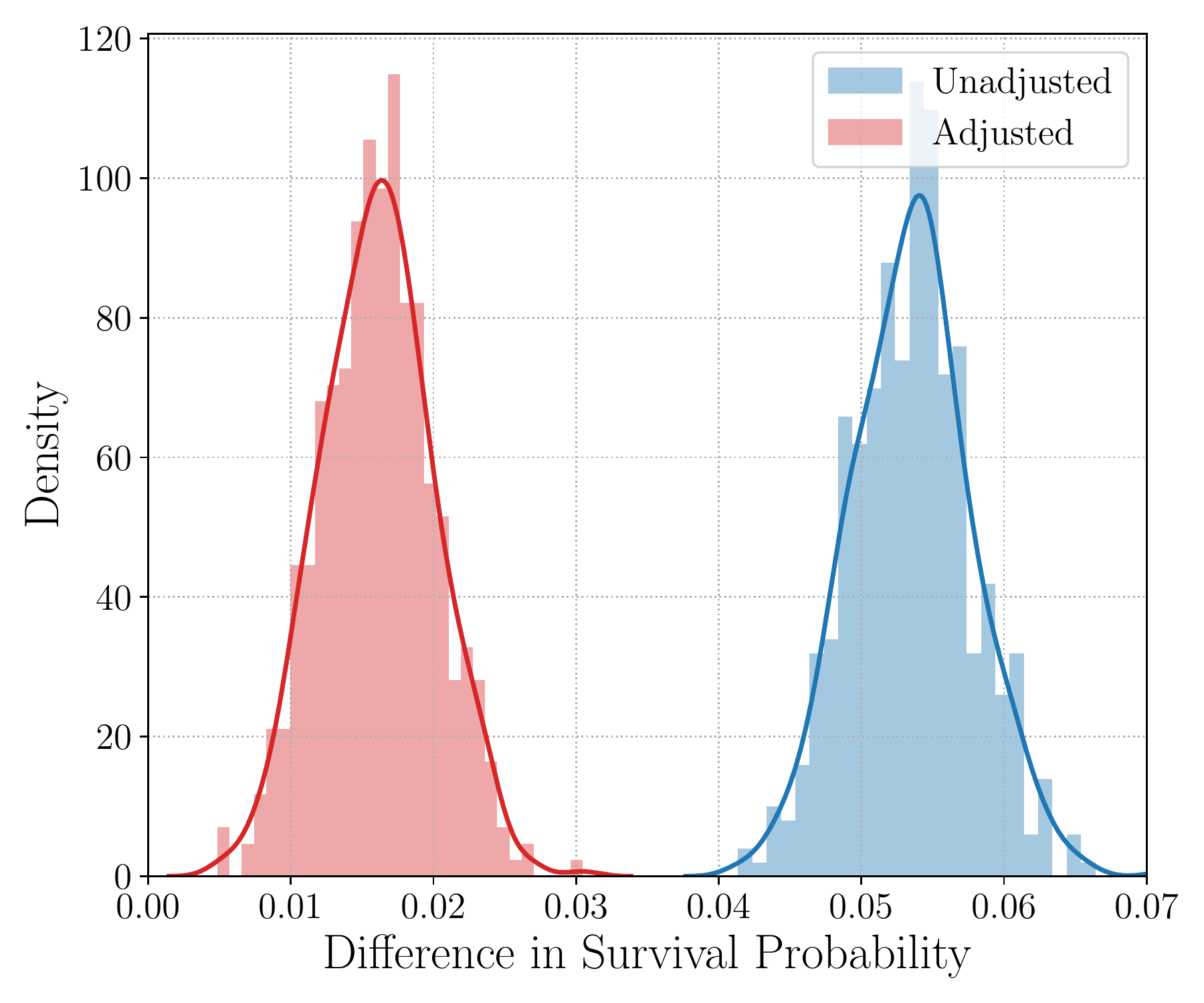}
    \subcaption{Risk Difference}
    \end{minipage}
    \begin{tabular}{r|c|c|c} \toprule
         \multirow{2}{*}{\textbf{Treatment Effect}} & \multicolumn{3}{c}{\textbf{10-year Event Horizon}} \\ \cline{2-4}
         {} & \textbf{Hazard Ratio} & \textbf{RMST} & \textbf{Risk Difference}  \\ \midrule
        Unadjusted & 0.728 $\pm$ 0.032 & 4.829 $\pm$ 0.684 & 0.054 $\pm$ 0.008  \\
        Adjusted &  0.890 $\pm$ 0.040 & 1.549 $\pm$ 0.618 & 0.016 $\pm$ 0.008 \\ \bottomrule
    \end{tabular}
    
    \caption{Estimated probability densities of the bootstrapped treatment effects to assess discrepancies in survival across regions. When bootstrapping is performed with weighted sampling based on region propensity to adjust for confounders, the discrepancies in survival between Greater California and Louisiana are much less pronounced. Bootstrap mean (standard deviation) results shown in the table summarize the observed mitigated difference.}
    \label{fig:seer-treatment-effects}
\end{figure}

A alternative method to assess effect size of the geographical region on survival involves comparing differences after adjusting for treatment propensity. We trained a Logistic Regression with an $\ell_2$ penalty by regression the Geographical Region on the set of confounding variables. The propensity scores were then employed as sampling weights for the treatment effects in terms of hazard ratios, restricted mean survival time (RMST), and risk difference as in \hyperref[fig:seer-treatment-effects]{Figure \ref*{fig:seer-treatment-effects}}. Adjusting for region propensity noticeably mitigates differences in treatment effects, indicating that mortality due to breast cancer is likely explained by confounding
socio-economic and physiological factors rather than solely geographic region.

\begin{lstlisting}
from auton_survival.metrics import treatment_effect
from sklearn.linear_model import LogisticRegression
# Create "treatment" labels
domain = regions=='Louisiana'
# Train a logistic regression model to compute "domain" propensity scores
model = LogisticRegression(penalty='l2').fit(features, domain)
propensity = model.predict_proba(features)[:, 1]
# Compute the treatment effect before and after adjusting for treatment propensity
effect = treatment_effect(metric='hazard_ratio', outcomes, domain,     
                             n_bootstrap=500)
adjusted_effect = treatment_effect(metric='hazard_ratio', outcomes, domain,                                            weights=propensity, n_bootstrap=500)
\end{lstlisting}

\subsection{Domain Adaptation}

\begin{minipage}{1\textwidth}
\centerline{\textbf{Area under ROC Curve}}\vspace{.5em}
\begin{minipage}{0.5\textwidth}
\centering
Louisiana $\rightarrow$ Greater California\\
\vspace{.5em}
\begin{tabular}{r|c|c|c} \toprule
    Horizon & 1-Year & 2-Year & 5-Year\\ \midrule
    Unadjusted & 0.8398 & 0.8177 & 0.8010 \\
    Adjusted & 0.8449 & 0.8220 & 0.8063 \\ \bottomrule
\end{tabular}
\end{minipage}%
\begin{minipage}{0.5\textwidth}
\centering
Greater California $\rightarrow$ Louisiana\\
\vspace{.5em}
\begin{tabular}{r|c|c|c} \toprule
    Horizon & 1-Year & 2-Year & 5-Year\\ \midrule
    Unadjusted & 0.8598 & 0.8343 & 0.8163 \\
    Adjusted & 0.8603 & 0.8348 & 0.8182 \\ \bottomrule
\end{tabular}
\end{minipage}\vspace{1em}
\centerline{\textbf{Brier Score}}\vspace{.5em}
\begin{minipage}{0.5\textwidth}
\centering
Louisiana $\rightarrow$ Greater California\\
\vspace{.5em}
\begin{tabular}{r|c|c|c} \toprule
    Horizon & 1-Year & 2-Year & 5-Year\\ \midrule
    Unadjusted & 0.0675 & 0.0967 & 0.1154 \\
    Adjusted & 0.0673 & 0.0963 & 0.1148 \\ \bottomrule
\end{tabular}
\end{minipage}%
\begin{minipage}{0.5\textwidth}
\centering
Greater California $\rightarrow$ Louisiana\\
\vspace{.5em}
\begin{tabular}{r|c|c|c} \toprule
    Horizon & 1-Year & 2-Year & 5-Year\\ \midrule
    Unadjusted & 0.0858 & 0.1119 & 0.1312 \\
    Adjusted & 0.0856 & 0.1118 & 0.1307 \\ \bottomrule
\end{tabular}
\end{minipage}
\captionof{table}{Domain Adaptation with importance weighting improves both calibration and discriminative performance when applying models trained on one domain to the other.}
\label{tab:domain-shift}
\end{minipage}

Consider a scenario when training data is available from Greater California (Lousiana) and is to be used to train models to estimate  risk for patients in Lousiana (Greater California). Discrepancies in the data distributions would naturally also translate into poorer generalization when using these models for individual level survival regression. 

It would thus help adjusting for these distributional differences using IWERM as introduced \hyperref[sec:iwerm]{Section \ref*{sec:iwerm}}. We train a logistic regression with an $\ell_2$ penalty to estimate the probability of an individual belonging to one of the two regions. The estimated probabilities are then used as importance weights to resample th individuals when training a Cox Model.

\hyperref[tab:domain-shift]{Table \ref*{tab:domain-shift}} present the performance of a linear Cox model in terms of discriminative performance (Area under ROC curve) and well as calibration (Brier Score) at predicting all cause mortality within 5, 10 and 15 years of entry into the study. Notice that the importance weighted Cox model has better discriminative performance and calibration as compared to the the unadjusted model.

\begin{lstlisting}
from auton_survival.estimators import SurvivalModel
from sklearn.linear_model import SGDClassifier
regions=['Greater California', 'Louisiana']
for source in regions:
    for target in regions:
        # Train a Logistic Regression to estimate treatment propensity scores
        domain_model = SGDClassifier(loss='log', penalty='l2', alpha=.1)
        domain_model.fit(features, regions==target)
        probs = domain_model.predict_proba(features[regions==source])[:, 1]
        # Use propensity scores to compute sample weights
        weights = ((probs)/(1-probs))**0.2
        # Fit DCPH model with weighted resampling
        model = SurvivalModel(model='dcph', layers=[100])
        model.fit(features[regions==source], outcomes[regions==source], 
                   weights, resample_size=10.0)
\end{lstlisting}

\section*{Conclusion}

We presented \package{}, an open source \texttt{Python} package encapsulating multiple pipelines to analyze censored time-to-event data ubiquitous in healthcare. %This functionality is especially applicable to healthcare where estimates of event risk conditioned by time can be use by clinicians to more appropriately triage patients and by healthcare institutions to optimize quality of care and cost.
Through the use of multiple code examples, notebooks, and illustrations we demonstrate the efficacy of \package{} to analyze complex healthcare data and answer clinical and epidemiological questions including real-world case studies. % involving studying geographical disparities in outcomes of breast cancer patients.}

{We acknowledge \package{} as one step towards promoting reproducible machine learning research for healthcare. We believe that the broader use of software packages such as ours will help accelerate and systematize impactful machine learning research for healthcare. We expect that ongoing and future collaborations with the machine learning and healthcare communities will allow us to further enhance the collection of open-source survival regression methodologies that can support reproducible analysis of censored time-to-event data.}

% \section*{Anonymity Statement}

% \textit{Our primary contribution is an open-source software package with an active user base making compliance with double blind peer review, difficult. We have made every possible genuine attempt at satisfying the requirements of double blind peer-review by anonymizing appropriate portions of our code repository, notebooks and documentation pages. We request the reviewers and area chairs to take a favorable view of this in the case that potentially author identifying information is inadvertently retained in the package.}
% \newpage
% \section{Case Study 2: ALLHAT/ACCORD?}

\section*{Acknowledgements}
The authors thank Xinyu (Rachel) Li, Vincent Jeanselme, Chufan Gao, Mononito Goswami, Roman Kauffman, Vedant Sanil, Shikha Reddy and Kishan Maharaj for their contributions. We thank the original developers of the \texttt{python} packages \texttt{scikit-learn}, \texttt{lifelines}, \texttt{scikit-survival}, \texttt{pycox} and \texttt{pytorch}, which \package{} is heavily influenced from.\\

\noindent This work was partially funded by DARPA under the award FA8750-17-2-0130.
\nocite{*}

{
\bibliography{ref}
\bibliographystyle{icml2020}
}

\appendix
\newpage

% \section{Right Censoring in Time-to-Event Prediction}
% \label{apx:censoring}

% \begin{figure}[!h]
%     \centering
%     \includegraphics[width=0.8\textwidth]{figures/censoring_fig.pdf}
%     \caption{\textbf{Censoring and Time-to-Event Predictions}: An illustration of Right Censoring. {\color{Maroon} \textbf{Patients A}} and {\color{Maroon} \textbf{C}} died {\color{Maroon}\textbf{1}} and {\color{Maroon}\textbf{4 years}} from entry into the study, whereas {\color{MidnightBlue} \textbf{Patients B}} and {\color{MidnightBlue}\textbf{D}} exited the study without experiencing death (were lost to follow up) at {\color{MidnightBlue}\textbf{2}} and {\color{MidnightBlue}\textbf{3 years}} from study entry.  \textit{Time-to-Event} or \textit{Survival Regression} thus involves adjusting estimates for such individuals whose outcomes were censored.}
%     \label{fig:censoring}
% \end{figure}

\section{Comparison to Other Packages}
\label{sec:comparison}

In the code snippets below we will compare the API of \package{} with the popular alternative, \texttt{pycox} to train a Deep Cox PH model on the SUPPORT dataset.

\begin{lstlisting}
from auton_survival import datasets, preprocessing 
# Load the SUPPORT Dataset
outcomes, features = datasets.load_dataset("SUPPORT")
# Preprocess (Impute and Scale) the features
features = preprocessing.Preprocessor().fit_transform(features)
\end{lstlisting}

\noindent Training a Deep Cox Proportional Hazards model with $\texttt{pycox}$
\begin{lstlisting}[escapeinside={(*}{*)}]
from torchtuples.tt.practical import MLPVanilla
from pycox.models import CoxPH
net = MLPVanilla(features.shape[1], [100], 1, output_bias=False) # Instantiate an MLP
model = CoxPH(net, tt.optim.Adam) # Instantiate a pycox model.
# Fit the pycox model.
model.fit(features, (outcomes.time, outcomes.event), batch_size=256, epochs=10)
\end{lstlisting}

\noindent Training a Deep Cox Proportional Hazards model with $\package{}$.
\begin{lstlisting}[escapeinside={(*}{*)}]
from auton_survival import models 
# Train a Deep Cox Proportional Hazards (DCPH) model
model = models.cph.DeepCoxPH(layers=[100])
model.fit(features, outcomes.time, outcomes.event, batch_size=256,)
# Predict risk at specific time horizons.
predictions = model.predict_risk(features, t=[8, 12, 16])
\end{lstlisting}

\begin{table}[!h]
\small
    \centering
    \begin{tabular}{rccc|c}
        & \multicolumn{4}{c}{} \\[2ex]
         \textbf{Utility} / \textbf{Package} & \rot{\texttt{lifelines}}  &  \rot{\texttt{pycox}} &  \rot{\texttt{scikit-survival}} & \rot{\package{}} \\ \midrule
        {Deep Survival Models} & \xmark & \cmark & \xmark & \cmark \\
        {Time-Varying Survival Analysis} & \cmark & \xmark & \xmark & \cmark\\
        {Counterfactual Estimation} & \xmark & \xmark & \xmark & \cmark\\
         {Subgroup Identification (Phenotyping)} & \xmark & \xmark & \xmark & \cmark \\
        {Treatment Effect Estimation} & \cmark & \xmark & \xmark & $\bigcirc$\\
        {Regression Metrics} & \cmark & \cmark & \cmark & $\bigcirc$\\
        {Cross-Validation} & \xmark & \xmark & \xmark & \cmark \\
        {Preprocessing} & \xmark & \cmark & \cmark & \cmark \\
         {Documentation and Examples}  & \cmark & \cmark & \cmark & \cmark \\ \midrule

    \end{tabular}
    \caption{{A comparison of \package{} to other open-source \texttt{python} packages for survival analysis. $\bigcirc$ indicates that the utility depends on one of other survival analysis packages.}}
    \label{tab:my_label}
\end{table}

\newpage

\newpage

\section{Additional details on the SEER case study}
\label{apx:seer-features}
\begin{table}[!h]
    \centering
    \begin{tabular}{r|l}
         \toprule
         \textbf{Name}& \textbf{Description}\\ \midrule
         \texttt{AGE}& {Age of participant}  \\
         \texttt{SEX}& {Sex of participant}  \\
         \texttt{RACE1V}&{Race/Ethnicity of participant}   \\
         \texttt{PRIMSITE}&{Site in which the primary tumor originated.}  \\ 
         \texttt{LATERAL}&{Side of a paired organ or body on which 
the tumor originated.}   \\
         \texttt{HISTO3V}&{Histologic Type ICD-O-3.}   \\
         \texttt{BEHO3V}&{Behavior Code ICD-O-3.}   \\
         \texttt{DX\_CONF}&{Diagnostic Confirmation.}   \\
         \texttt{SURGPRIF}&{Surgery of Primary Site.}   \\
         \texttt{SURGSITF}&{The surgical removal of distant 
tissue beyond the primary site.}   \\
         \texttt{NUMNODES}&{Number of Examined Nodes}   \\
         \texttt{NO\_SURG}&{Reason for no surgery.}   \\
         \texttt{SURGSITE}& {The removal of distant tissue beyond the primary site.}   \\
         \texttt{EOD10\_SZ}&{Tumor size.}   \\
         \texttt{EOD10\_EX}&{Tumor extension.}   \\
         \texttt{EOD10\_ND}&{The highest specific lymph node chain that is involved by the tumor.}   \\
         \texttt{EOD10\_PN}&{Regional nodes positive.}   \\
         \texttt{EOD10\_NE}&{regional nodes examined.}   \\
         \texttt{MALIGCOUNT}&{Total number of In Situ/malignant tumors for patient.}   \\
         \texttt{BENBORDCOUNT}&{Total number of benign/borderline tumors.}   \\
         \bottomrule

    \end{tabular}
    \caption{List of confounding features used for experiments involving the the SEER dataset.}
    \label{apx:seer-confounders}
\end{table}

%%%%%%%%%%%%%%%%%%%%%%%%%%%%%%%%%%%%%%%%%%%%%%%%%%%%%%%%%%%%%%%%%%%%%%%%%%%%%%%
%%%%%%%%%%%%%%%%%%%%%%%%%%%%%%%%%%%%%%%%%%%%%%%%%%%%%%%%%%%%%%%%%%%%%%%%%%%%%%%

\end{document}

%% file: math_commands.tex
\usepackage{amsmath,amsfonts,bm}

\def\eqref#1{equation~\ref{#1}}
\def\1{\bm{1}}

\def\va{{\bm{a}}}

\def\vt{{\bm{t}}}

\def\vx{{\bm{x}}}

\def\mS{{\bm{S}}}

\DeclareMathAlphabet{\mathsfit}{\encodingdefault}{\sfdefault}{m}{sl}
\SetMathAlphabet{\mathsfit}{bold}{\encodingdefault}{\sfdefault}{bx}{n}